\PassOptionsToPackage{table}{xcolor}
\documentclass[11pt,a4paper]{article}
\usepackage{times,latexsym}
\usepackage{url}
\usepackage[T1]{fontenc}
\usepackage[utf8]{inputenc}

\usepackage[acceptedWithA]{tacl2021v1}

\usepackage{xspace,mfirstuc,tabulary}

\newif\iftaclinstructions
\taclinstructionsfalse 
\iftaclinstructions
\renewcommand{\confidential}{}
\renewcommand{\anonsubtext}{(No author info supplied here, for consistency with
TACL-submission anonymization requirements)}
\newcommand{\instr}
\fi

\iftaclpubformat 

\else

\fi

\usepackage[main=english, arabic, russian]{babel}

\usepackage{multirow}
\usepackage{times}
\usepackage{latexsym}
\usepackage{microtype}
\usepackage{inconsolata}
\usepackage{graphicx}
\usepackage{import}
\usepackage{pgfplots}
\usepackage{csvsimple}
\usetikzlibrary{backgrounds}
\usepackage{tikz}
\usepackage{pgfplotstable}
\pgfplotsset{compat=1.18}
\usepgfplotslibrary{polar}
\usepackage{subcaption}
\usepackage{geometry}
\usepackage{etoolbox} 
\usetikzlibrary{positioning}
\usetikzlibrary{arrows}
\usepackage{float}
\usepackage{makecell}
\usepackage{hyperref}

\usepackage{amsmath}        

\usepackage{booktabs}
\usepackage[most]{tcolorbox}
\tcbset {
  base/.style={
    arc=0mm, 
    bottomtitle=0.5mm,
    boxrule=0mm,
    colbacktitle=black!10!white, 
    coltitle=black, 
    fonttitle=\bfseries, 
    left=2.5mm,
    leftrule=1mm,
    right=3.5mm,
    title={#1},
    toptitle=0.75mm, 
  }
}
\definecolor{brandblue}{rgb}{0.34, 0.7, 1}
\newtcolorbox{subbox}[1]{
  colframe=black!30!white,
  base={#1}
}
\usepackage{adjustbox}

\usepackage{highlight}
\usepackage[table]{xcolor}
\usepackage{pgffor}
\newcommand{\gblue}[1]{\gradientcell{#1}{0}{1}{white}{blue!60!gray}{100}}

\newcommand{\gdelta}[1]{%
  \ifdim #1 pt < 0pt
    \gradientcell{#1}{-1}{0}{red!60!gray}{white}{100}%
  \else
    \gradientcell{#1}{0}{1}{white}{green!60!gray}{100}%
  \fi
}

\newcommand{\impact}{\mbox{\textbf{IMPACT}}}

\newif\ifshowblue
\showbluetrue 

\newcommand{\blue}[1]{\ifshowblue\textcolor{black}{#1}\fi}

\newcommand{\correction}[1]{\textcolor{black}{#1}}

\definecolor{darkgreen}{rgb}{0.0, 0.5, 0.0}

\newcommand{\lucas}[1]{\textcolor{black}{#1}}

\title{\impact: Inflectional Morphology Probes Across Complex Typologies}

    \author{
     \textbf{Mohammed J. Saeed},
     \textbf{Tommi Vehvilainen},\\
     \textbf{Evgeny Fedoseev},
     \textbf{Sevil Caliskan},
     \textbf{Tatiana Vodolazova} 
\\
 \small{
   \{msaeed, tommi.vehvilainen, e\_fedoseev, scalskan, tvodolazova\}@apple.com
 }
\\
Apple
\\
}

\begin{document}

\maketitle

\begin{abstract}
Large Language Models (LLMs) have shown significant progress on various multilingual benchmarks and are increasingly used to generate and evaluate text in non-English languages. However, while they may produce fluent outputs, it remains unclear to what extent these models truly grasp the underlying linguistic complexity of those languages, particularly in morphology. To investigate this, we introduce IMPACT—a synthetically generated \lucas{evaluation framework} focused on inflectional morphology\lucas{, which we publicly release}—designed to evaluate LLM performance across five morphologically rich languages: Arabic, Russian, Finnish, Turkish, and Hebrew\correction{. IMPACT includes unit-test-style cases covering both shared and language-specific phenomena, from basic verb inflections (e.g., tense, number, gender) to unique features like Arabic’s reverse gender agreement and vowel harmony in Finnish and Turkish.}
\correction{We assess eight multilingual LLMs that, despite strong English performance, struggle with other languages and uncommon morphological patterns, especially when judging ungrammatical examples. We also show that Chain of Thought and Thinking Models can degrade performance. Our work exposes gaps in LLMs’ handling of linguistic complexity, pointing to clear room for improvement. \lucas{To support further research, we publicly release the IMPACT framework.}}

\end{abstract}

\section{Introduction}
\begin{figure}[t]
    \centering
    \includegraphics[width=7cm]{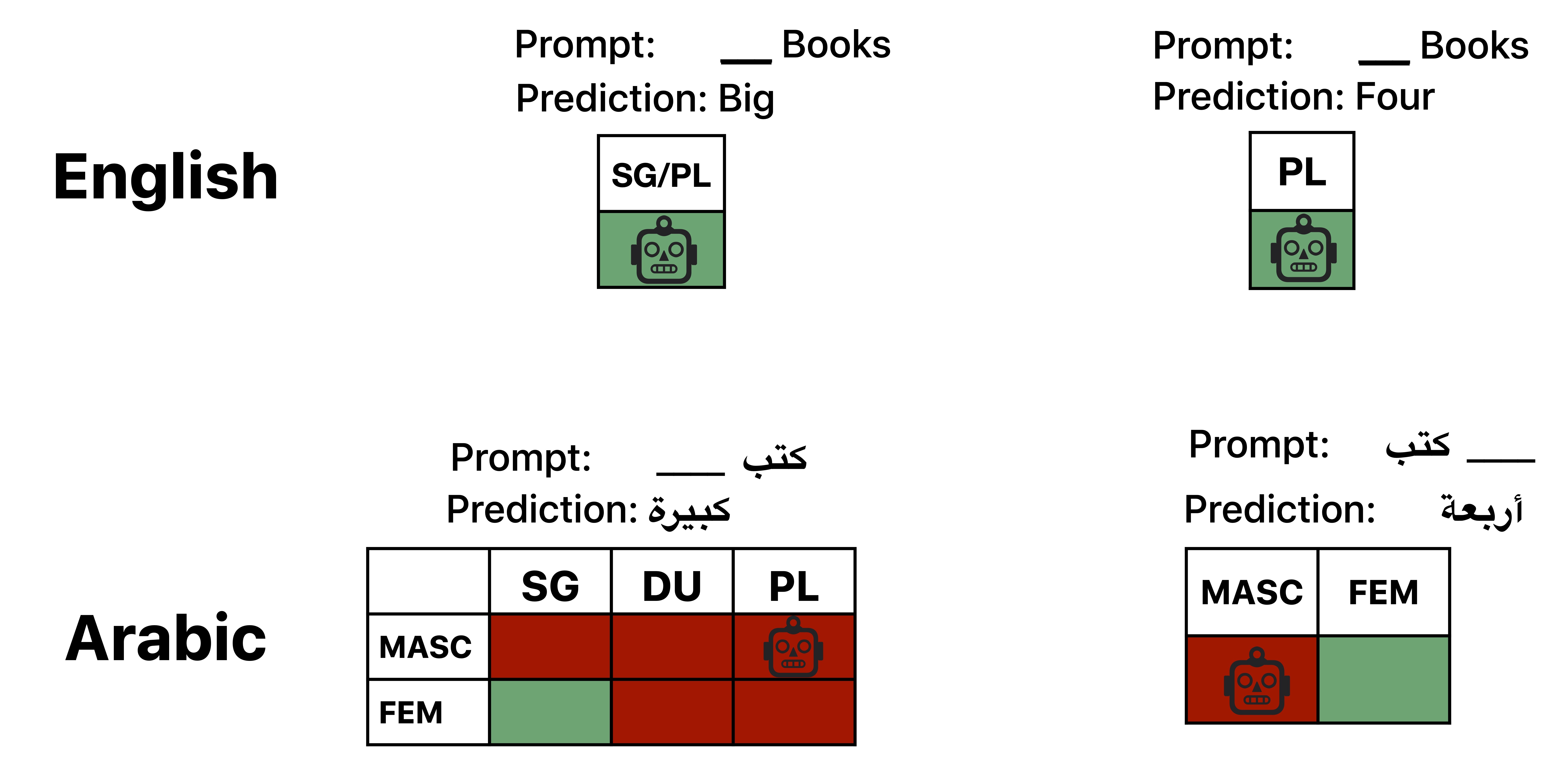}
    \caption{\correction{The top English examples show that (i) the adjective "big" remains unchanged for singular~\lucas{(SG)} and plural~\lucas{(PL)} nouns, and (ii) the number "four" inherently signals plurality. The language model (robot icon) correctly generates both. Below, the same examples in Arabic illustrate complex agreement with gender (MASC/FEM) and dual~\lucas{(DU)} forms. Green cells indicate correct feature matches; red cells show errors. On the left, the model wrongly uses a masculine plural adjective instead of the required singular feminine for plural inanimate nouns. On the right, it predicts masculine agreement for the cardinal number, whereas Arabic uses reverse gender agreement—feminine numerals with masculine nouns for 3 to 10.}}
    \label{fig:main_example}
\end{figure}
With over 7,000 known spoken languages worldwide, developing \correction{multilingual} language models has long been a central focus of Natural Language Processing (NLP) research. This interest has intensified with the advent of transformer-based models, such as mBERT~\citep{devlin-etal-2019-bert}, mT5~\citep{xue-etal-2021-mt5}, Orion-14B~\citep{Chen2024Orion14BOM}, and Aya~\citep{Ustun2024AyaMA}. These models are typically pre-trained on multilingual data~\citep{Scao2022BLOOMA1} to achieve strong performance across a broad range of languages. For example, \correction{Orion-14B's pre-training data is over 90\% English and Chinese. Japanese and Korean contribute more than 5\%, while the remaining portion includes a mix of languages such as Spanish, French, and Arabic.}
In addition, proprietary models like OpenAI's GPT-4 and Google's Gemini exhibit strong multilingual performance, setting high benchmarks in multilingual understanding and generation. For instance, GPT-4 shows outstanding performance on translated variants of MMLU~\citep{openai2024gpt4technicalreport}. Similarly, Gemini Ultra excels on very low-resource languages~\citep{geminiteam2025geminifamilyhighlycapable}.

\begin{figure}[t]
    \centering
    \includegraphics[width=8cm]{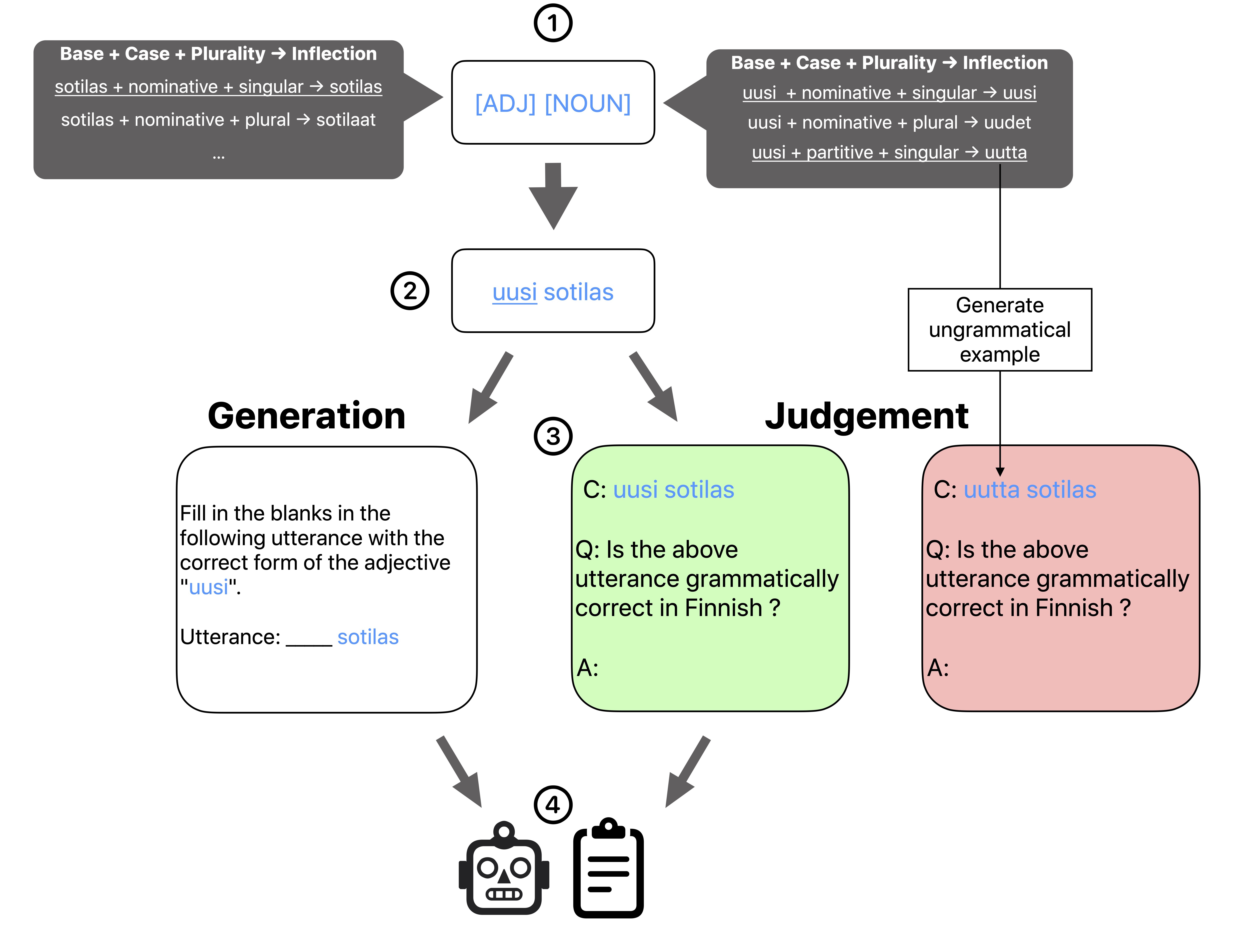}
    \caption{\correction{IMPACT workflow: (1) Identify templates and prepare placeholder data (Section~\ref{sec:3-1}); (2) generate utterances with correct inflections using morphological agreement logic (Section~\ref{sec:3-2}); (3) create prompts for \texttt{Generation} and \texttt{Judgement} tasks (Section~\ref{sec:3-3}), including ungrammatical examples by mismatching inflections; (4) evaluate LLM performance.}}
    \label{fig:impact}
\end{figure}

While large language models (LLMs) consistently show improvements on various multilingual benchmarks, their performance is often evaluated through downstream tasks—such as translation and question answering—that may overlook finer linguistic intricacies. Specifically, it is uncertain whether LLMs effectively capture linguistic features of languages with rich morphological structures \correction{(Figure~\ref{fig:main_example})}. For instance, Finnish, an agglutinative language with 14 grammatical cases, constitutes only about 1.35\% of the mC4 dataset~\citep{xue-etal-2021-mt5}, raising concerns about its adequate representation during model training. 
To address this, we take a step back and systematically evaluate LLMs from a morphological perspective. We introduce \impact, a collection of tests designed to assess whether LLMs' \correction{grasp of inflectional morphology} 
across languages (Figure~\ref{fig:impact}). For each language, we (i) identify templates \correction{targeting} specific morphological aspects, (ii) generate \correction{controlled} utterances, and (iii) test LLMs in two scenarios: one where the LLM predicts the correct inflection (\texttt{Generation}), and another \correction{popular scenario} where the LLM \correction{judges} grammatical and ungrammatical utterances (\texttt{Judgement}). 

\correction{We analyze five morphologically rich languages—Arabic, Hebrew, Russian, Turkish, and Finnish—focusing on inflectional morphology, with derivational morphology left for future work. Finnish and Turkish are agglutinative, using linear affixation for single grammatical functions, while the others are fusional, combining multiple functions within single affixes. Each language also presents unique features: Arabic includes dual forms and reverse gender agreement; Finnish and Turkish lack grammatical gender; and Russian distinguishes three genders.}

Similar to unit tests, which ensure that a system behaves as expected given specific inputs, \impact{} generates "morphological unit tests" \correction{to evaluate LLMs’ handling of inflectional morphology}. \correction{These targeted tests help uncover blind spots and limitations in model performance.} Our contributions are as follows:

\begin{itemize}
    \item We introduce \impact, an \lucas{evaluation framework} of carefully constructed tests that evaluate LLMs on morphologically rich languages. It covers both shared and language-specific features, including around 33K test instances across various languages (Section~\ref{sec:3}).
        
    \item We assess models in two distinct scenarios: \texttt{Generation}, where the LLM predicts the correct inflection, and \texttt{Judgement}, where the LLM evaluates the grammaticality of provided utterances. We show how LLMs are good judges for grammatical examples but lag behind when judging ungrammatical examples, with generation accuracy falling in between (Section~\ref{subsec:4-2}).
            
    \item We study, in Section~\ref{subsec:4-3}, that the impact of Chain-of-Thought (CoT) and "Thinking Modes" is inconsistent—often beneficial but sometimes detrimental.
    
    \item We qualitatively analyze failing tests across LLMs and languages in Section~\ref{subsec:4-4}, revealing that LLMs can identify correct morphological rules but often fail to apply them reliably. 

\end{itemize}

\section{Related Work \& Background}
\noindent\textbf{Multi-linguality.} Unlike monolingual LLMs that focus on a single language~\citep{devlin-etal-2019-bert, martin-etal-2020-camembert}, multilingual LLMs handle multiple languages—usually centered on English—but face the "curse of multi-linguality" 
~\citep{chang-etal-2024-multilinguality}.
Their training spans several phases including multilingual pre-training~\citep{blevins-etal-2024-breaking}, and \correction{non-English} instruction fine-tuning~\citep{holmstrom-doostmohammadi-2023-making}.
This work focuses on instruction-tuned multilingual LLMs~\citep{shaham-etal-2024-multilingual}.

\noindent\textbf{LLM-As-a-Judge.} The LLM-as-a-Judge framework~\citep{chiang-lee-2023-large, zheng2023judging} has gained traction by automating evaluations, with LLMs replacing human experts. Despite its growing adoption, \correction{it suffers from} variability and inconsistency in judgement~\citep{bavaresco2024judge}.
In multilingual contexts, GPT-4 \correction{tends to give overly positive scores}, and \correction{unreliable quality} metrics~\citep{hada-etal-2024-large},  \correction{pointing to clear room for improvement in multilingual LLM evaluation~\citep{son2025mmevalmultilingualmetaevaluationbenchmark}.}
\correction{Despite their fluency, LLMs may not truly understand what they generate—a disconnect known as the Generative AI Paradox~\citep{west2024the}. Most prior studies focus on English; our work broadens this to other languages.}

\noindent\textbf{Morphological Probing.} \correction{Behavioral testing frameworks like CheckList\citep{ribeiro-etal-2020-beyond} reveal model weaknesses overlooked by standard metrics. 
In multilingual contexts}, M2C~\cite{hlavnova-ruder-2023-empowering} probes LLMs on linguistic features across 12 diverse languages, while TEA~\citep{k-etal-2022-multilingual} automatically extracts templates from machine-translated instances of source-language templates in a cost-effective manner. Complementing these approaches, wug tests~\citep{anh-etal-2024-morphology} are used to assess language models’ morphological generalization.
While these works share a similar spirit to ours, 
our templates are designed to capture language-specific aspects of inflectional morphology, with input from native speakers to ensure linguistic accuracy and relevance. We also expand the evaluation to two scenarios: (i) \texttt{Generation}, \correction{requiring correct inflection}, and (ii) \texttt{Judgement}, where the LLM \correction{assesses grammatical correctness of both well-formed and ungrammatical examples.}

\noindent\textbf{Morphologically Rich Languages.} \correction{These} languages feature extensive inflectional morphology, where words change form to express grammatical features such as tense, case, gender, number, and person. 
Further details are provided in Section~\ref{sec:3-2}.
\begin{table*}[t]
    \centering
    \resizebox{0.7\textwidth}{!}{
    \begin{tabular}{l|ccc||ccc||ccc||ccc}
    \toprule
  & \multicolumn{3}{c}{GF2} & \multicolumn{3}{c}{Aya} & \multicolumn{3}{c}{EuroLLM} & \multicolumn{3}{c}{Qwen3} \\

 & Gen & JY & JN & Gen & JY & JN & Gen & JY & JN & Gen & JY & JN \\
\midrule
eng-com-1 & \gblue{.985} & \gblue{1.} & \gblue{1.} & \gblue{.889} & \gblue{.967} & \gblue{.916} & \gblue{1.} & \gblue{.979} & \gblue{.519} & \gblue{1.} & \gblue{1.} & \gblue{1.} \\
eng-com-2 & \gblue{1.} & \gblue{1.} & N/A & \gblue{.880} & \gblue{.731} & N/A & \gblue{1.} & \gblue{.980} & N/A & \gblue{1.} & \gblue{1.} & N/A \\
eng-com-3 & \gblue{1.} & \gblue{.990} & N/A & \gblue{.990} & \gblue{.807} & N/A & \gblue{.880} & \gblue{.900} & N/A & \gblue{1.} & \gblue{.980} & N/A \\
\midrule
ara-com-1 & \gblue{.893} & \gblue{.972} & \gblue{.915} & \gblue{.360} & \gblue{.925} & \gblue{.821} & \gblue{.175} & \gblue{.975} & \gblue{0} & \gblue{.625} & \gblue{.830} & \gblue{.705} \\
ara-com-2 & \gblue{.928} & \gblue{.986} & \gblue{.837} & \gblue{.397} & \gblue{.667} & \gblue{.776} & \gblue{0} & \gblue{.990} & \gblue{0} & \gblue{.303} & \gblue{.753} & \gblue{.549} \\
ara-com-3 & \gblue{.951} & \gblue{.889} & \gblue{.887} & \gblue{.559} & \gblue{.886} & \gblue{.708} & \gblue{.296} & \gblue{.897} & \gblue{.136} & \gblue{0} & \gblue{.576} & \gblue{.802} \\
ara-1 & \gblue{.500} & \gblue{.555} & \gblue{.343} & \gblue{.713} & \gblue{.800} & \gblue{.533} & \gblue{.417} & \gblue{.850} & \gblue{.152} & \gblue{.649} & \gblue{.713} & \gblue{.738} \\
\midrule
heb-com-1 & \gblue{.961} & \gblue{.935} & \gblue{.954} & \gblue{.606} & \gblue{.929} & \gblue{.891} & \gblue{0} & \gblue{.878} & \gblue{.065} & \gblue{.612} & \gblue{.841} & \gblue{.807} \\
heb-com-2 & \gblue{.896} & \gblue{.761} & \gblue{.891} & \gblue{.345} & \gblue{.319} & \gblue{.942} & \gblue{.160} & \gblue{.955} & \gblue{0} & \gblue{.518} & \gblue{.770} & \gblue{.656} \\
heb-com-3 & \gblue{.990} & \gblue{.992} & \gblue{.959} & \gblue{.919} & \gblue{.874} & \gblue{.947} & \gblue{.611} & \gblue{.835} & \gblue{.208} & \gblue{.749} & \gblue{.924} & \gblue{.922} \\
heb-1 & \gblue{.669} & \gblue{.723} & \gblue{.617} & \gblue{.719} & \gblue{.949} & \gblue{.611} & \gblue{.141} & \gblue{.916} & \gblue{.137} & \gblue{.457} & \gblue{.838} & \gblue{.684} \\
\midrule
rus-com-1 & \gblue{.990} & \gblue{1.} & \gblue{.999} & \gblue{.851} & \gblue{.990} & \gblue{.948} & \gblue{.801} & \gblue{.985} & \gblue{.119} & \gblue{.898} & \gblue{.982} & \gblue{.986} \\
rus-com-2 & \gblue{.940} & \gblue{.990} & \gblue{.778} & \gblue{.818} & \gblue{.950} & \gblue{.582} & \gblue{.854} & \gblue{1.} & \gblue{0} & \gblue{.827} & \gblue{.980} & \gblue{.875} \\
rus-com-3 & \gblue{.967} & \gblue{.973} & \gblue{0} & \gblue{.356} & \gblue{.898} & \gblue{.786} & \gblue{0} & \gblue{.972} & \gblue{0} & \gblue{.694} & \gblue{.942} & \gblue{.927} \\
rus-1 & \gblue{.789} & \gblue{.711} & \gblue{.246} & \gblue{.458} & \gblue{.789} & \gblue{.192} & \gblue{.429} & \gblue{.990} & \gblue{.034} & \gblue{.359} & \gblue{.597} & \gblue{.420} \\
\midrule
tur-com-1 & \gblue{.964} & \gblue{.990} & \gblue{.683} & \gblue{.183} & \gblue{.953} & \gblue{.436} & \gblue{.484} & \gblue{.949} & \gblue{0} & \gblue{.446} & \gblue{.970} & \gblue{.798} \\
tur-com-2 & \gblue{.925} & \gblue{.990} & \gblue{.898} & \gblue{.847} & \gblue{.909} & \gblue{.670} & \gblue{.873} & \gblue{.990} & \gblue{0} & \gblue{.723} & \gblue{.980} & \gblue{.587} \\
tur-com-3 & \gblue{1.} & \gblue{.985} & \gblue{.882} & \gblue{.863} & \gblue{.698} & \gblue{.827} & \gblue{.763} & \gblue{.889} & \gblue{.294} & \gblue{.611} & \gblue{.959} & \gblue{.857} \\
tur-1 & \gblue{1.} & \gblue{1.} & \gblue{.471} & \gblue{.941} & \gblue{.882} & \gblue{.529} & \gblue{.353} & \gblue{1.} & \gblue{.294} & \gblue{.824} & \gblue{1.} & \gblue{.471} \\
tur-2 & \gblue{.940} & \gblue{.936} & \gblue{.643} & \gblue{.035} & \gblue{.659} & \gblue{.586} & \gblue{0} & \gblue{1.} & \gblue{0} & \gblue{.178} & \gblue{.738} & \gblue{.635} \\

\midrule
fin-com-1 & \gblue{1.} & \gblue{1.} & \gblue{.980} & \gblue{.558} & \gblue{.940} & \gblue{.899} & \gblue{.788} & \gblue{1.} & \gblue{.149} & \gblue{.772} & \gblue{.990} & \gblue{.956} \\
fin-com-2 & \gblue{.950} & \gblue{1.} & \gblue{.889} & \gblue{.509} & \gblue{.836} & \gblue{.463} & \gblue{.836} & \gblue{1.} & \gblue{.039} & \gblue{.549} & \gblue{1.} & \gblue{.750} \\
fin-com-3 & \gblue{.919} & \gblue{.980} & \gblue{.941} & \gblue{0} & \gblue{.683} & \gblue{.662} & \gblue{.142} & \gblue{.951} & \gblue{0} & \gblue{.399} & \gblue{.851} & \gblue{.833} \\
fin-1 & \gblue{.105} & \gblue{1.} & \gblue{0} & \gblue{0} & \gblue{.828} & \gblue{0} & \gblue{.585} & \gblue{1.} & \gblue{0} & \gblue{0} & \gblue{.622} & \gblue{.320} \\
fin-2 & \gblue{.911} & \gblue{.978} & \gblue{.489} & \gblue{0} & \gblue{.200} & \gblue{.578} & \gblue{.200} & \gblue{1.} & \gblue{0} & \gblue{.178} & \gblue{.689} & \gblue{.533} \\
\bottomrule
\end{tabular}
}
\caption{The scores for Generation (Gen) and Judgement for positive (JY) and negative (JN) instances for each LLM on each template. The darker the color (the higher the score), the better. \lucas{LLM} performance degrades for JN while they perform better for JY and Gen somewhere in between. Smaller LLMs perform poorly for JN.}
\label{tab:temp_agg_mini}
\end{table*}

\section{Templates}
\label{sec:3}
We begin by introducing the templates used to evaluate various morphological features across languages (Section~\ref{sec:3-1}). Based on these templates, we generate a wide range of utterances and evaluate LLMs in two settings: \texttt{Generation} and \texttt{Judgement}, as detailed in Section~\ref{sec:3-2}. The final prompts , which are passed to the LLMs, are discussed in Section~\ref{sec:3-3}.

\subsection{Identifying Templates} 
\label{sec:3-1}
The goal of a template is to serve as a general blueprint for generating utterances to evaluate linguistic phenomena. \correction{For example, Template com-1 (Table~\ref{tab:comm}, Appendix~\ref{sec:appendixB}) contains placeholders like \texttt{[NAME]} and \texttt{[VERB]} to test subject-verb agreement. The \texttt{[NAME]} placeholder, drawn from a wordlist, varies by gender, case, and number, creating multiple combinations. Each unique placeholder combination, called an "Evaluation Unit", is the smallest evaluated element (see Table~\ref{tab:eu_example}, Appendix~\ref{sec:appendixD}). }
Placeholders are filled \correction{with} language-specific word lists \correction{using UniMorph~\citep{batsuren-etal-2022-unimorph} for inflected forms}. While many templates are shared across languages—despite differences in morphological agreement rules—some languages require their own templates to accommodate unique grammatical structures (see Table~\ref{tab:specific_templates}\correction{, Appendix~\ref{sec:appendixB})}. We provide further details in the following subsection.
\subsection{Template Logic}
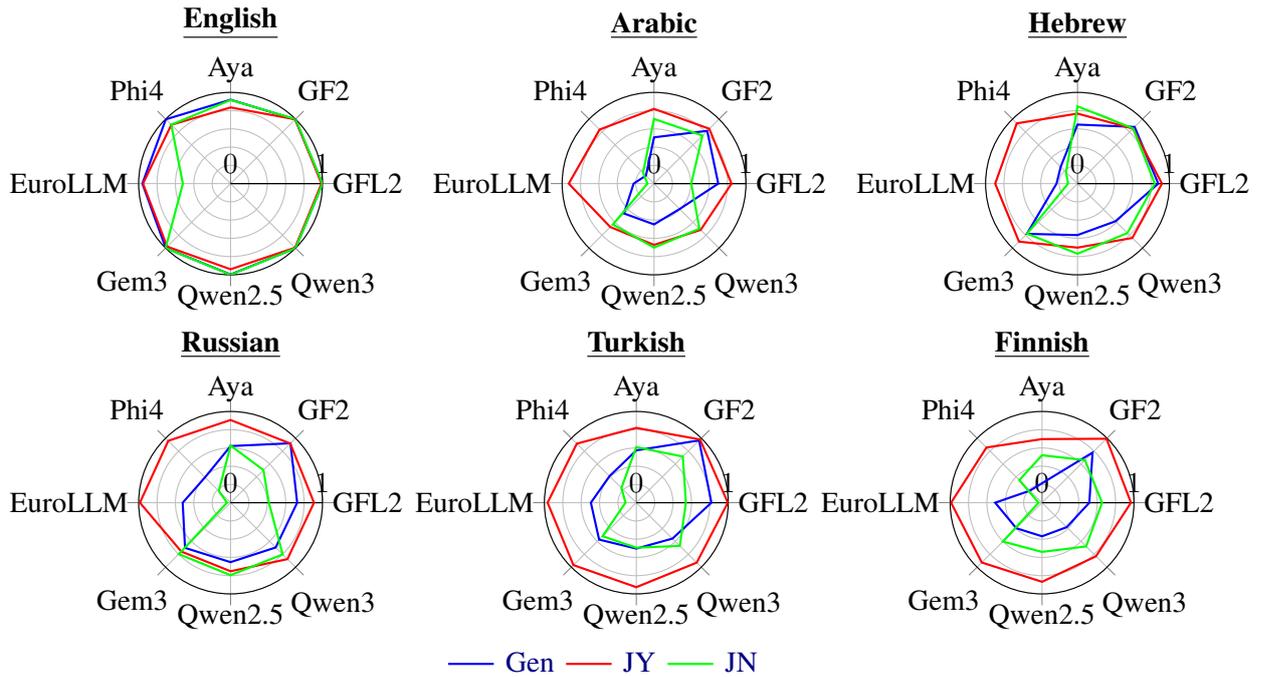
\begin{figure*}[t]
\centering

\newtoggle{firstplot}
\toggletrue{firstplot}

\foreach \lang/\language in {eng/English, ara/Arabic, heb/Hebrew, rus/Russian, tur/Turkish,  fin/Finnish}{
  \begin{subfigure}{0.3\textwidth}
    \centering
    \begin{tikzpicture}
\begin{polaraxis}[
    title={\underline{\textbf{\language}}},
    xtick={0,45,90,135,180,225,270,315},
    xticklabels={GFL2, GF2, Aya, Phi4, EuroLLM, Gem3, Qwen2.5, Qwen3},
    ymin=0, ymax=1,
    ytick={0,1},
    minor y tick num=4,
    minor y tick style={draw=none},
    grid=both,
    width=4cm,
    height=4cm,
    legend to name=commonlegend,
    legend columns=3,
    legend style={at={(0.5,1)}, anchor=south, draw=none, column sep=2pt}
]
    \addplot [mark=none, draw=blue, thick]
        table[x=angle, y=Gen, col sep=comma] {radarData/\lang.csv} -- cycle;
    \addplot [mark=none, draw=red, thick]
        table[x=angle, y=JY, col sep=comma] {radarData/\lang.csv} -- cycle;
    \addplot [mark=none, draw=green, thick]
        table[x=angle, y=JN, col sep=comma] {radarData/\lang.csv} -- cycle;

    \iftoggle{firstplot}{
        \addlegendentry{Gen}
        \addlegendentry{JY}
        \addlegendentry{JN}
        \togglefalse{firstplot}
    }{}
    \end{polaraxis}
    \end{tikzpicture}
  \end{subfigure}
  \hfill
}

\ref{commonlegend}

\caption{Radar plots showing Gen, JY, and JN scores for each language. \lucas{We observe that LLMs judge overall  positive instances better than negative ones.}}
\label{fig:radar}
\end{figure*}

\label{sec:3-2}
We now discuss the logic behind the templates used, starting with the common templates \correction{(Table~\ref{tab:comm})} and followed by the language-specific ones\correction{(Table~\ref{tab:specific_templates})}. While some languages have specific templates, even the common ones differ in their underlying logic. In addition to the morphologically rich languages, we apply the common templates on English as a sanity check.

\noindent\textbf{com-1.} This template evaluates agreement between names and verbs. The \texttt{[VERB]} appears in the indicative mood, either in the present or past tense. \texttt{[NAME]} is drawn from a list of names, with the option to combine multiple names using conjunctions. In Arabic and Hebrew, \texttt{[NAME]} and \texttt{[VERB]} must agree in both gender and plurality, with a dedicated inflection for the dual form. In Finnish and Turkish, only plurality is relevant for agreement. In Russian, agreement is required in plurality, and gender is additionally marked when the verb is in the past tense.
\\
\noindent\textbf{com-2.} This template is similar to \textbf{com-1}, but \texttt{[VERB]} is in the imperative mood. For Arabic and Hebrew, \texttt{[NAME]} and \texttt{[VERB]} must agree on gender and plurality, whereas in Russian, Finnish and Turkish, only plurality is required for agreement.
\\
\noindent\textbf{com-3.} This template measures the agreement between adjectives and nouns. \texttt{[NOUN]} is populated from a list of occupations (e.g., doctor, engineer, teacher) and objects (e.g., book, pen). Animacy distinction is crucial here. For Arabic, \texttt{[NOUN]} and \texttt{[ADJ]} must agree in gender and plurality, with a special rule for plural inanimate nouns, where the adjective must be singular and feminine. This distinction also helps assess neuter objects in Russian, which does not apply to templates \textbf{com-1} and \textbf{com-2} as they address humans. In Russian, agreement is required in plurality and case, as well as gender when the \texttt{[NOUN]} is singular. In Finnish, agreement occurs in case and plurality. In Turkish, no agreement is required, as adjectives remain singular regardless of the noun’s plurality.
\\
\noindent\textbf{ara-1\&heb-1} In Arabic and Hebrew, when the cardinal number ranges from 3 to 10, it exhibits reverse gender agreement with the noun it modifies—that is, a masculine noun takes a feminine numeral and vice versa. This template tests whether this is correctly applied.
\\
\noindent\textbf{rus-1.} Russian employs motion verbs that distinguish between unidirectional forms—used for single, goal-oriented actions often in progress—and multidirectional forms, which describe habitual, repeated, or general movement. This template tests whether the appropriate verb form is chosen based on context, such as \foreignlanguage{russian}{идти} (to go on foot, unidirectional) versus \foreignlanguage{russian}{ходить} (to go on foot habitually, multidirectional). Only the present tense is targeted.
\\
\correction{\noindent\textbf{tur-1.} In Turkish, locative case suffixes follow vowel and consonant harmony rules. This template tests the correct selection of the suffix vowel and consonant based on the phonological properties of the noun’s final sound.}
\\
\correction{\noindent\textbf{tur-2.} In Turkish, the suffix -miş marks indirect evidentiality, showing that the speaker infers the event rather than witnessing it directly. This template tests the correct use of this suffix to convey indirect evidence.}
\\
\correction{\noindent\textbf{fin-1.} In Finnish, case inflection for city names is lexically determined. This template tests the correct use of the illative versus the allative case, depending on the specific city. For example, Helsinki takes the illative form Helsinkiin, while Tampere uses the allative form Tampereelle.}\\
\correction{\noindent\textbf{fin-2.} Finnish vowel harmony constrains how words are inflected based on the presence of front, back, or neutral vowels. This template tests whether the elative case is formed in accordance with these harmony patterns.}
\subsection{Generating Data}
\begin{table}[t]
\centering
\resizebox{\columnwidth}{!}{
\begin{tabular}{l|cccccccc}
\toprule
\textbf{Model} & GFL2 & GF2 & Aya & Phi4 & EuroLLM & Gem3 & Qwen2.5 & Qwen3 \\
\midrule
\textbf{MS} & 0.688 & \textbf{0.820} & 0.539 & 0.269 & 0.167 & 0.625 & 0.574 & 0.603 \\
\bottomrule
\end{tabular}
}
\caption{Model Score (MS) for each model.}
\label{tab:model_ms_scores}
\end{table}
\label{sec:3-3}
Once the utterances are synthesized, we can begin preparing the prompts for both scenarios: \texttt{Generation} and \texttt{Judgement}. Full prompt examples for both scenarios, including system instructions, are provided in Appendix~\ref{sec:prompts}.

\noindent\textbf{Generation.} Given an utterance, the objective is to assess whether the LLM can predict the correct inflection. This scenario takes the form of a "Fill in the blanks" exercise, where one inflection is omitted from each generated utterance, and the LLM is tasked with predicting the correct inflection based on the base form of the word to be filled in. An example is shown in Figure~\ref{fig:impact}. 
For certain generation scenarios, additional context is provided in the prompt. In the case of \textbf{rus-1}, we specify that the verb should sound natural to a native speaker, as speakers sometimes mix both unidirectional and multidirectional forms. For \textbf{tur-1}, we indicate that the inflection should be in the locative case. In \textbf{tur-2}, we specify that the information was obtained indirectly, conveying evidentiality. For \textbf{com-1}, the tense (present or past) is clarified. For \textbf{com-2}, we specify that the utterance is in the imperative mood. Specifically for Russian, we note that we are in an informal setting, since the plural verb form can agree with a singular noun in a formal context. For \textbf{com-3}, adjectives are specified to be in the base form, without comparative or superlative inflections.

\noindent\textbf{Judgement.} Given a grammatically correct utterance, the goal here is to evaluate whether the LLM can judge the grammatical correctness of the utterance. Once the grammatically correct form is generated, it is perturbed to create a negative example. In most cases, this perturbation involves changing the inflection so that it no longer respects the morphological agreement. For templates that assess vowel harmony (such as \textbf{fin-2} and \textbf{tur-1}), the negative example is created by altering the suffix to break the vowel harmony. We prompt the LLM to output either "Yes" or "No" in response to whether the utterance is grammatically correct. We evaluate both positive and negative utterance separately. 

\noindent\textbf{Sampling.} As the number of examples grows significantly with the number of evaluation units, we sample 50 prompts per evaluation unit for each scenario. For instance, when evaluating a unit that combines a singular masculine noun with a plural masculine verb, we generate 50 distinct prompts and compute the success rate. However, in some languages and templates, the number of possible evaluation units can be very large. In template \textbf{com-1}, Russian yields 311 distinct combinations in the \texttt{Judgement} scenario due to the interplay of plurality, gender, and case—resulting in a high number of negative examples. To manage this, we down-sample the negative examples to match the number of positive examples when the total number of evaluation units exceeds 100. This ensures that each template is evaluated across all its valid combinations while maintaining computational feasibility. 
\begin{table*}[t]
    \centering
    \resizebox{0.7\textwidth}{!}{
    \begin{tabular}{l|ccc||ccc||ccc||ccc}
\toprule
  & \multicolumn{3}{c}{GF2} & \multicolumn{3}{c}{Aya} & \multicolumn{3}{c}{EuroLLM} & \multicolumn{3}{c}{Qwen3} \\
 & Gen & JY & JN & Gen & JY & JN & Gen & JY & JN & Gen & JY & JN \\
\midrule
eng-com-1 & \gdelta{-.015} & \gdelta{0} & \gdelta{0} & \gdelta{-.106} & \gdelta{-.033} & \gdelta{-.084} & \gdelta{.036} & \gdelta{-.021} & \gdelta{-.461} & \gdelta{.020} & \gdelta{.016} & \gdelta{0} \\
eng-com-2 & \gdelta{0} & \gdelta{.020} & N/A & \gdelta{-.120} & \gdelta{.439} & N/A & \gdelta{0} & \gdelta{-.020} & N/A & \gdelta{0} & \gdelta{.095} & N/A \\
eng-com-3 & \gdelta{.010} & \gdelta{.041} & N/A & \gdelta{.142} & \gdelta{.162} & N/A & \gdelta{-.120} & \gdelta{-.100} & N/A & \gdelta{0} & \gdelta{.467} & N/A \\
\midrule
ara-com-1 & \gdelta{.893} & \gdelta{.270} & \gdelta{.114} & \gdelta{.360} & \gdelta{.222} & \gdelta{.002} & \gdelta{.175} & \gdelta{.033} & \gdelta{0} & \gdelta{.625} & \gdelta{.320} & \gdelta{-.016} \\
ara-com-2 & \gdelta{.005} & \gdelta{.097} & \gdelta{.581} & \gdelta{.397} & \gdelta{.305} & \gdelta{-.105} & \gdelta{0} & \gdelta{.007} & \gdelta{0} & \gdelta{.303} & \gdelta{.115} & \gdelta{.197} \\
ara-com-3 & \gdelta{.015} & \gdelta{.113} & \gdelta{.045} & \gdelta{-.189} & \gdelta{.105} & \gdelta{-.081} & \gdelta{-.037} & \gdelta{-.044} & \gdelta{.136} & \gdelta{-.532} & \gdelta{-.091} & \gdelta{-.034} \\
ara-1 & \gdelta{-.449} & \gdelta{.063} & \gdelta{-.254} & \gdelta{.006} & \gdelta{.025} & \gdelta{.365} & \gdelta{.175} & \gdelta{-.010} & \gdelta{.038} & \gdelta{-.090} & \gdelta{.075} & \gdelta{.318} \\
\midrule
heb-com-1 & \gdelta{.078} & \gdelta{.077} & \gdelta{.077} & \gdelta{-.208} & \gdelta{.006} & \gdelta{.240} & \gdelta{0} & \gdelta{-.122} & \gdelta{.065} & \gdelta{.315} & \gdelta{.209} & \gdelta{-.091} \\
heb-com-2 & \gdelta{-.027} & \gdelta{.337} & \gdelta{.039} & \gdelta{.345} & \gdelta{.049} & \gdelta{-.021} & \gdelta{.160} & \gdelta{-.045} & \gdelta{0} & \gdelta{-.030} & \gdelta{.325} & \gdelta{-.155} \\
heb-com-3 & \gdelta{-.005} & \gdelta{.151} & \gdelta{-.018} & \gdelta{.035} & \gdelta{.094} & \gdelta{-.028} & \gdelta{.241} & \gdelta{-.165} & \gdelta{.208} & \gdelta{.118} & \gdelta{.185} & \gdelta{-.007} \\
heb-1 & \gdelta{-.311} & \gdelta{-.097} & \gdelta{.057} & \gdelta{.195} & \gdelta{-.041} & \gdelta{.453} & \gdelta{-.135} & \gdelta{-.084} & \gdelta{.137} & \gdelta{-.054} & \gdelta{-.015} & \gdelta{.378} \\
\midrule
rus-com-1 & \gdelta{.020} & \gdelta{0} & \gdelta{.002} & \gdelta{-.026} & \gdelta{.028} & \gdelta{-.052} & \gdelta{.006} & \gdelta{-.015} & \gdelta{-.675} & \gdelta{.028} & \gdelta{.107} & \gdelta{-.014} \\
rus-com-2 & \gdelta{.117} & \gdelta{-.010} & \gdelta{.619} & \gdelta{-.036} & \gdelta{-.010} & \gdelta{.236} & \gdelta{-.004} & \gdelta{0} & \gdelta{0} & \gdelta{.035} & \gdelta{.122} & \gdelta{.035} \\
rus-com-3 & \gdelta{-.014} & \gdelta{.150} & \gdelta{-.992} & \gdelta{.356} & \gdelta{.630} & \gdelta{-.207} & \gdelta{-.180} & \gdelta{-.024} & \gdelta{0} & \gdelta{.520} & \gdelta{.442} & \gdelta{-.036} \\
rus-1 & \gdelta{-.075} & \gdelta{.049} & \gdelta{.133} & \gdelta{-.402} & \gdelta{.015} & \gdelta{.057} & \gdelta{-.009} & \gdelta{-.010} & \gdelta{.034} & \gdelta{.217} & \gdelta{.046} & \gdelta{.015} \\
\midrule
tur-com-1 & \gdelta{.027} & \gdelta{.037} & \gdelta{.395} & \gdelta{-.324} & \gdelta{.095} & \gdelta{.303} & \gdelta{.119} & \gdelta{-.041} & \gdelta{0} & \gdelta{-.165} & \gdelta{.051} & \gdelta{.313} \\
tur-com-2 & \gdelta{-.055} & \gdelta{-.010} & \gdelta{.418} & \gdelta{.023} & \gdelta{.004} & \gdelta{.438} & \gdelta{.033} & \gdelta{-.010} & \gdelta{0} & \gdelta{-.052} & \gdelta{.143} & \gdelta{-.055} \\
tur-com-3 & \gdelta{.020} & \gdelta{.056} & \gdelta{.009} & \gdelta{-.101} & \gdelta{-.071} & \gdelta{-.158} & \gdelta{-.175} & \gdelta{-.106} & \gdelta{.139} & \gdelta{-.379} & \gdelta{.238} & \gdelta{-.097} \\
tur-1 & \gdelta{.176} & \gdelta{0} & \gdelta{.177} & \gdelta{.588} & \gdelta{.058} & \gdelta{-.059} & \gdelta{.177} & \gdelta{0} & \gdelta{.294} & \gdelta{0} & \gdelta{0} & \gdelta{.118} \\
tur-2 & \gdelta{.090} & \gdelta{-.054} & \gdelta{.007} & \gdelta{-.102} & \gdelta{.606} & \gdelta{-.400} & \gdelta{-.033} & \gdelta{.767} & \gdelta{-.747} & \gdelta{.058} & \gdelta{.175} & \gdelta{-.097} \\
\midrule
fin-com-1 & \gdelta{.005} & \gdelta{.005} & \gdelta{.007} & \gdelta{-.136} & \gdelta{.167} & \gdelta{.274} & \gdelta{-.136} & \gdelta{0} & \gdelta{.149} & \gdelta{0} & \gdelta{.048} & \gdelta{.001} \\
fin-com-2 & \gdelta{-.019} & \gdelta{0} & \gdelta{.556} & \gdelta{.168} & \gdelta{-.154} & \gdelta{.463} & \gdelta{.379} & \gdelta{0} & \gdelta{.039} & \gdelta{.150} & \gdelta{.031} & \gdelta{.445} \\
fin-com-3 & \gdelta{.008} & \gdelta{.023} & \gdelta{.365} & \gdelta{0} & \gdelta{-.031} & \gdelta{.383} & \gdelta{-.171} & \gdelta{-.047} & \gdelta{0} & \gdelta{.399} & \gdelta{-.036} & \gdelta{.013} \\
fin-1 & \gdelta{.105} & \gdelta{0} & \gdelta{0} & \gdelta{-.564} & \gdelta{-.142} & \gdelta{0} & \gdelta{-.318} & \gdelta{0} & \gdelta{0} & \gdelta{0} & \gdelta{.432} & \gdelta{-.618} \\
fin-2 & \gdelta{.133} & \gdelta{.045} & \gdelta{-.400} & \gdelta{-.333} & \gdelta{-.667} & \gdelta{.511} & \gdelta{-.511} & \gdelta{0} & \gdelta{0} & \gdelta{-.489} & \gdelta{.067} & \gdelta{-.245} \\
\bottomrule
\end{tabular}

}

\caption{Score differences for Generation (Gen), and Judgements of positive (JY) and negative (JN) instances for some LLMs when shifting from Direct prompting to Chain-of-Thought (CoT). \lucas{CoT} does not always help, and sometimes hurts performance.}
\label{tab:diff_temp_agg_mini}
\end{table*}

\section{Experiments \& Results}
We first start by introducing the experimental setup in Section~\ref{subsec:4-1}. We then assess quantitatively how LLMs perform in both \texttt{Generation} and \texttt{Judgement} in Section~\ref{subsec:4-2}, followed by an analysis of the impact of CoT and Thinking Models (Section~\ref{subsec:4-3}). Finally, a qualitative analysis of the failing evaluation units is presented in Section~\ref{subsec:4-4}.
\subsection{Experimental Setup}
\label{subsec:4-1}
\correction{\noindent\textbf{Setup.} We evaluate LLM performance across languages in both \texttt{Generation} and \texttt{Judgement} scenarios. In the \texttt{Generation} scenario, a test is considered successful if the correct inflection appears in the generated output; invalid responses fail. In the \texttt{Judgement} scenario, a test is deemed successful if the output contains either "Yes" or "No". In case of an invalid response, we label the example as invalid and randomly sample a response following ~\citet{bavaresco2024judge}. The valid response rate of each model can be found in Table~\ref{tab:llms} (Appendix~\ref{sec:appendixC}). Models were prompted with English chain-of-thought reasoning~\citep{huang-etal-2023-languages, ShiSF0SVCTRZ0W23}. No few-shot examples were provided, as our goal was to evaluate the models' inherent capabilities in a zero-shot setting~\citep{hlavnova-ruder-2023-empowering,efrat-etal-2023-lmentry}.} \lucas{Examples for the used prompts are found in Section~\ref{subsec:cot_prompts} in Appendix~\ref{sec:prompts}.}
\\
\noindent\textbf{Models.} We evaluate several proprietary and open-source LLMs, as listed in Table~\ref{tab:llms}. All open-source models are deployed using vLLM~\citep{kwon2023efficient}. Decoding is performed with the temperature parameter set to 0. For Qwen3, we \lucas{follow best practices by avoiding} greedy decoding, and instead use the recommended parameters for the "Non-Thinking" mode~\citep{yang2025qwen3technicalreport}. 

\subsection{Can LLMs Effectively Generate and Judge Grammaticality?}
\label{subsec:4-2}
 We begin our analysis by evaluating LLMs’ \texttt{Generation} and \texttt{Judgement} capabilities for each template, reporting separate scores for each. For \texttt{Judgement}, we report results on grammatical and ungrammatical sentences independently. 

Since each template consists of multiple evaluation units, we aggregate the unit-level scores into a single template score using the harmonic mean. The harmonic mean is chosen to heavily penalize poorly performing evaluation units, preventing strong results in some units from masking weaknesses in others. 
\lucas{In contrast, the arithmetic mean fails to account for such disparities, and the geometric mean is less stringent in  penalization.} More formally, for language 
$lang$:
 $S^{\text{lang}} = \mathrm{HM}\bigl(\mathrm{acc}_{\mathrm{EU}_i}^{\text{lang}}\bigr), \quad \text{for } S \in \{\text{Gen}, \text{JY}, \text{JN}\}$
 where $\mathrm{acc}_{\mathrm{EU}_i}^{\text{lang}}$ denotes the accuracy of the ith evaluation unit $EU_i$ and $HM$ is the harmonic mean. We report \correction{partial} \texttt{Generation} scores (Gen) alongside \texttt{Judgement} scores for grammatical (JY) and ungrammatical (JN) utterances in Table~\ref{tab:temp_agg_mini}. \blue{Full results are shown in Table~\ref{tab:temp_agg} in Appendix~\ref{sec:appendixE}.}

\noindent\textbf{English.} LLMs perform consistently well on all English templates across both \texttt{Generation} and \texttt{Judgement} tasks. This is expected, as English presents relatively low morphological complexity compared to other languages. In some templates, negative instances are entirely absent, making the task even less challenging for the models. EuroLLM struggles with ungrammatical utterances. 

\noindent\textbf{Arabic.} \lucas{With the exception of GF2, LLMs generally struggle with both Gen and JN across most templates.} While models are more successful at identifying grammatical constructions (as reflected in higher JY scores), their ability to generate or reject incorrect forms remains limited. This pattern is even more pronounced in language-specific templates involving reverse gender agreement.

\noindent\textbf{Hebrew.} Most LLMs perform well, especially in JY, with varying results across Gen and JN. This asymmetry between JY and JN is also reflected in \textbf{heb-1}.

\noindent\textbf{Russian.} Most LLMs achieve relatively strong performance on Gen and JY across common templates. However, JN proves more difficult—especially in the \textbf{rus-com-3} template—where several models struggle to reject ungrammatical forms. These limitations are also highlighted in the language-specific template \textbf{rus-1}.

\noindent\textbf{Turkish.} We observe a similar trend where JN consistently lags behind both Gen and JY, indicating that models find it harder to identify ungrammatical constructions than to produce or recognize grammatical ones. This issue becomes more pronounced in language-specific tasks involving evidentiality and vowel harmony suffixes.

\noindent\textbf{Finnish.} Models show relatively stronger performance in JY, indicating an ability to recognize well-formed constructions. However, Gen and JN scores remain low, suggesting difficulties in both producing and rejecting morphologically complex or malformed inputs. These challenges are amplified in language-specific templates that involve lexical casing and vowel harmony.\\
\noindent\textbf{Language.} To analyze performance across languages, we aggregate scores over all templates per language using the arithmetic mean. We choose the arithmetic mean instead of the harmonic mean to avoid issues with zero values. Figure~\ref{fig:radar} presents these language-level scores, showing how LLMs underperform in negative judgements.\\
\noindent\textbf{Model.} We then compute an overall model score, $MS = HM(Gen, Jud)$, combining \texttt{Generation} and \texttt{Judgement} scores via harmonic mean, where $Jud = HM(JY, JN)$ computes a single score for \texttt{Judgement}. As shown in Table~\ref{tab:model_ms_scores}, GF2 achieves the highest overall score, while smaller models like Phi4 and EuroLLM consistently underperform—likely due to shallow pattern-matching rather than deeper grammatical understanding.

In conclusion, LLMs excel at recognizing grammatical sentences, struggle more with ungrammatical ones, and perform moderately on generation. While larger models generally outperform smaller ones, they still falter in complex scenarios—especially language-specific templates involving non-trivial morphological agreement patterns.
\begin{table}[t]
    \centering
    \resizebox{\columnwidth}{!}{
    \begin{tabular}{l|ccc||ccc||ccc}
    \toprule
     & \multicolumn{3}{c}{GF2} & \multicolumn{3}{c}{EuroLLM} & \multicolumn{3}{c}{Qwen3} \\
     & Gen & JY & JN & Gen & JY & JN & Gen & JY & JN \\
     &  &  &  &  &  &  &  &  &  \\
    \midrule
    eng-com-1 &  &  &  &  &  & X &  &  &  \\
    eng-com-2 &  &  &  &  &  &  &  &  &  \\
    eng-com-3 &  &  &  &  &  &  &  &  &  \\
    \midrule
    ara-com-1 &  &  & X & X &  & X & X &  & X \\
    ara-com-2 &  &  & X & X &  & X & X & X & X \\
    ara-com-3 &  & X & X & X &  & X & X & X & X \\
    ara-1 & X & X & X & X &  & X &  &  &  \\
    \midrule
    heb-com-1 &  &  &  & X &  & X & X &  & X \\
    heb-com-2 &  & X &  & X &  & X & X &  & X \\
    heb-com-3 &  &  &  & X &  & X & X &  &  \\
    heb-1 &  &  & X & X &  & X & X &  & X \\
    \midrule

    rus-com-1 &  &  &  &  &  & X &  &  &  \\
    rus-com-2 &  &  &  &  &  & X &  &  &  \\
    rus-com-3 &  &  & X & X &  & X & X &  & X \\
    rus-1 &  & X & X & X &  & X & X & X & X \\
    \midrule
    tur-com-1 &  &  & X & X &  & X & X &  &  \\
    tur-com-2 &  &  &  &  &  & X & X &  & X \\
    tur-com-3 &  &  &  &  &  & X & X &  &  \\
    tur-1 &  &  & X & X &  & X &  &  & X \\
    tur-2 &  &  & X & X &  & X & X & X & X \\
    \midrule

    fin-com-1 &  &  &  &  &  & X &  &  &  \\
    fin-com-2 &  &  &  &  &  & X & X &  &  \\
    fin-com-3 &  &  & X & X &  & X & X &  & X \\
    fin-1 & X &  & X & X &  & X & X & X & X \\
    fin-2 &  &  & X & X &  & X & X &  &  \\

    \bottomrule
    \end{tabular}}
    \caption{Checklist of LLMs that fail on specific templates. A failure is marked with an "X". An LLM is considered to fail a template if at least one of its evaluation units exhibits a failure rate of 60\% or higher.}
    \label{tab:checklist_mini}
\end{table}

\subsection{Does "Thinking" Help?}
\label{subsec:4-3}
We assess the impact of (i) CoT and (ii) "Thinking Mode" on model performance, and more specifically, on \texttt{Generation} and \texttt{Judgement} capabilities.\\
\\
\noindent\textbf{CoT.} We re-run the same experiments as before, but \textbf{without} CoT, where the LLM is prompted to directly output the inflected form for the generation scenario and either Yes or No for the \texttt{Judgement} scenario.
We then report $\Delta S=S_{CoT}-S_{Direct}$ for $S \in \{Gen,JY,JN,Jud,MS\}$. \correction{Partial} template-level results are shown in Table~\ref{tab:diff_temp_agg_mini}. \blue{\blue{Full results are shown in Table~\ref{tab:diff_temp_agg} in Appendix~\ref{sec:appendixE}.}}
Overall, CoT generally improves performance for LLMs, though its effects are highly variable. For example, CoT boosted the JY score of Qwen2.5 on \textbf{tur-2}, but simultaneously reduced the JN score for the same template. Analyzing the reasoning chains, we found that Qwen2.5 correctly identifies that the verb should appear in reported speech but fails to verify this properly. This pattern is also evident in the Gen scores for \textbf{fin-1} and \textbf{fin-2} with GFL2, where longer reasoning chains introduce more potential points of failure—either by applying incorrect grammatical rules or generating incorrect inflections.
English templates generally perform well. CoT has little effect on GF2, boosts Phi4 and Qwen2.5, but lowers EuroLLM’s performance.
Finally, no model showed improvement from CoT in JN for \textbf{rus-com-3}. Analysis of reasoning chains reveals that models consistently failed to detect grammatical errors involving case and plurality, often misclassifying incorrect examples as correct. Notably, Gem3 and Qwen2.5 were exceptions, but they were unable to make correct judgements.

\noindent\textbf{Thinking Mode.}
\correction{Recent advances in test-time scaling enable models to dynamically adjust their reasoning within a fixed "thinking budget," enhancing performance on complex tasks without additional training~\citep{snell2025scaling}.}
We evaluate how Qwen3's "Thinking Mode" compares to the "Non-Thinking" mode (Direct). Due to the high computational cost of these experiments, we restrict our analysis to a single model, Qwen3. We set the maximum number of tokens to 32K and adjust the temperature, top-k, and top-p parameters according to the recommended values in~\cite{yang2025qwen3technicalreport}. Following the previous section, we report the delta scores. \blue{Results are presented at the template level in Table~\ref{tab:think_diff_temp_agg} (Appendix~\ref{sec:appendixE}) and at the language level in Table~\ref{tab:think_diff_lang_agg} (Appendix~\ref{sec:appendixF}).}
Similar to CoT, we observe that the impact is highly variable, boosting performance on some specific templates \lucas{while} reducing it on others with more gains on Arabic, Russian, and Hebrew. Turkish and Finnish show varying improvements with a strong loss for JN for \textbf{fin-1}.

In conclusion, our observations align with recent findings~\citep{SpragueYRJWSZYM25, ma2025reasoningmodelseffectivethinking}, indicating that CoT and "Thinking Mode" do not consistently enhance the performance of LLMs and may introduce new errors, highlighting the need for more robust reasoning strategies.

\subsection{Where do LLMs Fail?}
\label{subsec:4-4}
Our previous analysis focused on aggregated metrics; we now examine performance at the evaluation unit level to identify units with significant failures. We define failing units as those with an accuracy below 60\% (failure rate $\geq$ 40\%). Table~\ref{tab:checklist_mini} lists templates where at least one evaluation unit failed for \correction{a few models} and scenarios, marked by "X". \blue{Full results are shown in Table~\ref{tab:checklist} (Appendix~\ref{sec:appendixK}).} We then concentrate on the best-performing model, GF2 with CoT. Using this threshold, we analyze 29 evaluation units—about 41\% from Arabic, 21\% from Finnish, and roughly 72\% related to negative judgements (JN). We then manually review all failing examples, examining their reasoning chains to identify common failure patterns.

\noindent\textbf{Arabic.}  In \textbf{com-3}, GF2 often violates gender agreement with dual plurality, mislabeling incorrect examples as correct. It also struggles with plural inanimate nouns, confusing singular feminine adjectives with typical feminine plural agreement, and sometimes misclassifies grammatical sentences as ungrammatical. For cardinal number agreement requiring reverse gender, GF2 fails in all judgement units except masculine nouns, generating regular gender agreement instead. Reasoning chains show GF2 recognizes the need for reverse agreement but fails to produce the correct cardinal number. In \textbf{com-1}, GF2 incorrectly accepts masculine dual verbs with feminine dual subjects and masculine plural verbs with dual nouns. Similarly, in \textbf{com-2}, it wrongly judges plural verbs with dual nouns as correct.
\\
\noindent\textbf{Hebrew.} For \textbf{com-2}, GF2 sometimes flags plural feminine nouns with matching plural feminine verbs as ungrammatical, often due to the verb’s archaic form. Similar to Arabic, in \textbf{heb-1}, GF2 recognizes the need for reverse agreement but fails to correctly judge ungrammatical utterances.
\\
\noindent\textbf{Russian.} For \textbf{com-3}, failures mainly stem from difficulty in identifying correct case markings, leading to misjudged ungrammatical utterances. In \textbf{rus-1}, GF2 incorrectly accepts the unidirectional form of a motion verb when paired with habitual motion. \\
\noindent\textbf{Turkish.} For \textbf{tur-2}, GF2 judges sentences as correct if the reported tense is accurate, even when verb–noun plurality mismatches occur. A similar issue arises in \textbf{com-1}. In \textbf{tur-1}, GF2 fails to detect ungrammatical vowel harmony errors.\\
\noindent\textbf{Finnish.}  For \textbf{fin-1}, GF2 fails to correctly identify any ungrammatical utterances, particularly struggling with the use of illative and allative cases for city names. This difficulty also affects the \texttt{Generation} phase. In contrast, for \textbf{fin-2}, GF2 generally reasons well about ungrammatical sentences but struggles during the final verification step involving the elative case. At this stage, GF2 sometimes incorrectly produces the correct inflection by assuming it is part of the input, resulting in the output being mistakenly labeled as grammatical. \\
\noindent\textbf{Common Failures.} We also identified evaluation units where all models consistently failed. For Arabic, all models failed to (i) correctly judge that feminine plural inanimate nouns do not agree with feminine plural adjectives, and (ii) recognize that masculine dual nouns do not agree with masculine plural verbs in both imperative and indicative contexts. \correction{For Finnish, all models failed on template \textbf{fin-1} when the task involved generating the correct allative case inflection, and also failed to properly reject ungrammatical examples.} For Russian, all models struggled to judge sentences requiring a multidirectional motion verb instead of the unidirectional form. Finally, for Turkish, all models failed on (i) \textbf{tur-1} by accepting forms with incorrect vowel harmony, and (ii) \textbf{tur-2} by incorrectly judging plurality agreement.





\section{Conclusion}

\correction{We present \impact, a framework for evaluating LLMs' morphological abilities across diverse languages. Testing both shared and language-specific grammar in generation and grammatical judgement reveals persistent struggles with morphologically rich languages. Models judge grammatical sentences well but falter on ungrammatical ones, with generation performance in between. Prompting strategies like Chain-of-Thought and Thinking Modes help but can introduce new errors.}

\blue{\section*{Limitations}
While our work highlights the limitations of LLMs in morphologically rich languages, several constraints remain. First, we focus mainly on inflectional morphology, leaving other crucial phenomena—such as derivational morphology, negation, and telicity marking—unaddressed. Second, template creation demands considerable linguistic expertise and relies on the Unimorph dataset, which is not fully human-verified; although we performed manual checks on some examples to ensure quality, some instances may still be wrong, limiting scalability. Third, our analysis targets utterances rather than fully composed semantic sentences, and many morphological patterns cannot be easily captured by templated constructions, which challenges broader generalization. Additionally, our reliance on accuracy as the primary metric may overlook performance subtleties, especially given the potential for unfaithful reasoning traces. Finally, due to computational cost constraints, (i) our evaluation of "Thinking Mode" is limited to a single model—Qwen3—and (ii) we perform sampling at the evaluation unit level and test on a carefully curated suite, rather than conducting multiple trials and reporting means and standard deviations.}

\blue{\section*{Acknowledgments}
The authors would like to thank Eero Vitie for fruitful discussions regarding the Finnish templates and their evaluation, and Ilay Sofer for verifying Hebrew data.}


\clearpage
\appendix
\section{Prompt Examples}
\label{sec:appendixA}
\label{sec:prompts}
\subsection{CoT System Prompts}
\label{subsec:cot_prompts}
\begin{subbox}{Generation}
  You are an expert in \texttt{English}.
Give step by step reasoning before you answer, and when you’re ready to answer, please use the format "Final Answer: ANSWER" where ANSWER is the inflected word.
  \tcblower
  Fill in the blanks in  the following utterance with the correct form of the \texttt{verb}  \texttt{"eat"}. Make sure that the verb is in the past tense. \\\\
Utterance: \texttt{Mohammed} \underline{\hspace{1cm}}
\end{subbox}

\begin{subbox}{Judgement}
  You are an expert in \texttt{Arabic}.
When you’re ready to answer, please use the format "Final Answer: ANSWER" where ANSWER is either "Yes" or "No".
  \tcblower
Given the following utterance, determine if it is grammatically correct or not in Modern Standard Arabic. \\
Utterance: \textRL{مدير طويل}\\
Answer: 
\end{subbox}

\subsection{Direct System Prompts}
\label{subsec:direct_prompts}

\begin{subbox}{Generation}
  You are an expert in \texttt{English}.
  Provide the inflected form only, without any explanation. 
  \tcblower
Fill in the blanks in  the following utterance with the correct form of the \texttt{verb}  \texttt{"eat"}. Make sure that the verb is in the past tense. \\
Utterance: \texttt{Mohammed} \underline{\hspace{1cm}}
\end{subbox}

\begin{subbox}{Judgement}
  You are an expert in \texttt{Arabic}.
Answer with Yes or No only. Do not provide any explanations.
  \tcblower
Given the following utterance, determine if it is grammatically correct or not in Modern Standard Arabic. \\
Utterance: \textRL{مدير طويل}\\
Answer: 
\end{subbox}


\section{Common and Language-Specific Templates}
\label{sec:appendixB}
Common templates between languages are found in Table~\ref{tab:comm}, whereas Table~\ref{tab:specific_templates} shows specific language templates. More details can be found in Section~\ref{sec:3-2}.
\begin{table}[H]
\centering
\begin{tabular}{l|lp{2cm}}
\toprule
\textbf{id} & \textbf{Template}     & \textbf{Example}\\
\toprule
com-1       &  {[}NAME{]} {[}VERB{]} & Tom eats. \\ 
\midrule
com-2       & Hey {[}NAME{]}, {[}VERB{]}!  & Hey Tom and Beth, return! \\ 
\midrule

com-3       & [ADJ] [NOUN]  & big houses \\
\bottomrule
\end{tabular}
\caption{Common Templates}
\label{tab:comm}
\end{table}
\begin{table}[H]
\centering
\begin{tabular}{l|p{4cm}p{2cm}}
\toprule
\textbf{id} & \textbf{Template}     & \textbf{Example}\\
\toprule
ara-1       &  {[}NUM{]} {[}NOUN{]} & \textRL{عشرة راكبون} \\ 
\midrule
heb-1       &  {[}NUM{]} {[}NOUN{]} &
\includegraphics[width=2cm]{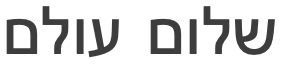}
\\ 
\midrule
rus-1       & \foreignlanguage{russian}{Я} [MOTION\_VERB] [PREPOSITION] [PLACE] [TIME].  & \foreignlanguage{russian}{Я иду в магазин сейчас.} \\
\midrule

tur-1       & [VERB]yim/yım.   & Yuvadayım. \\

tur-2       & [NAME] [VERB]  & Nazlı savunmuş. \\
\midrule
fin-1       & Aloitetaan reittiohjeet [CITY].  & Aloitetaan reittiohjeet Espooseen. \\ 

fin-2       & Pidän [PLACE].  & Pidän kaupungista.\\
\midrule

\bottomrule
\end{tabular}
\caption{Language-Specific Templates}
\label{tab:specific_templates}
\end{table}

\section{Models}
\label{sec:appendixC}
Table~\ref{tab:llms} represents the Valid Response Rate (VRR) per model for the \texttt{Judge} scenario using CoT. The \texttt{Generation} scenario did not have any invalid responses.
We observe a relatively lower VRR for \textbf{Qwen3} using the "Non-Thinking" Mode compared to other LLMs, due to lengthy responses, which might be motivated by the  thinking-mode-centric architecture and training of Qwen3. The "Thinking Mode" with a higher max number of tokens yielded a VRR of 0.9998.
\begin{table}[H]
\centering
\resizebox{\columnwidth}{!}{
\begin{tabular}{llc}
\toprule
\textbf{Model Name} & \textbf{Alias} & \textbf{VRR} \\
\midrule
\href{https://huggingface.co/google/gemma-3-27b-it}{google/gemma-3-27b-it} & Gem3 &0.9707 \\
\href{https://huggingface.co/CohereLabs/aya-expanse-32b}{CohereLabs/aya-expanse-32b} & Aya &0.9960 \\
\href{https://huggingface.co/Qwen/Qwen2.5-32B-Instruct}{Qwen/Qwen2.5-32B-Instruct} & Qwen2.5 & 0.9997 \\
\href{https://huggingface.co/Qwen/Qwen3-32B}{Qwen/Qwen3-32B} & Qwen3 & 0.9190 \\
\href{https://huggingface.co/utter-project/EuroLLM-9B-Instruct}{utter-project/EuroLLM-9B-Instruct} & EuroLLM & 0.9655 \\
\href{https://huggingface.co/microsoft/Phi-4-mini-instruct}{microsoft/Phi-4-mini-instruct} & Phi4 & 0.9996 \\
\href{https://cloud.google.com/vertex-ai/generative-ai/docs/models/gemini/2-0-flash-lite}{gemini-2.0-flash-lite} & GFL2 & 0.9999 \\
\href{https://cloud.google.com/vertex-ai/generative-ai/docs/models/gemini/2-0-flash}{gemini-2.0-flash} & GF2 & 0.9995 \\
\bottomrule
\end{tabular}}
\caption{LLMs used with their aliases and valid response rate.}
\label{tab:llms}
\end{table}

\section{Evaluation Unit Example}
\label{sec:appendixD}
Table~\ref{tab:eu_example} shows the accuracy of GF2 on a few evaluation units for the template \textbf{ara-com-2} in the \texttt{Judge} scenario. 
Nouns and verbs can either be masculine (MASC) or feminine (FEM) and either singular (SG), dual (DU), or plural (PL). The verbs are in the imperative (IMP) mood. An evaluation unit is a concatenation of all these morphosyntactic features. In the \texttt{Judge} scenario, we append either "Yes" or "No" to indicate whether the examples are grammatical or not, and to differentiate it from the evaluation units of the \texttt{Generation} scenario.
\begin{table}[H]
    \centering
\resizebox{\columnwidth}{!}{
\begin{tabular}{l|c}
\toprule
 Evaluation Unit & Accuracy \\
\midrule
noun-MASC-DU-verb-MASC-PL-IMP-No & {.300} \\
noun-FEM-DU-verb-FEM-PL-IMP-No & {.500} \\
noun-FEM-PL-verb-MASC-PL-IMP-No & {.700} \\
noun-FEM-DU-verb-MASC-PL-IMP-No & {.760} \\
noun-FEM-SG-verb-MASC-SG-IMP-No & {.860} \\
noun-FEM-SG-verb-FEM-PL-IMP-No & {.900} \\
noun-FEM-DU-verb-FEM-SG-IMP-No & {.940} \\
noun-FEM-PL-verb-MASC-DU-IMP-No & {.980} \\
noun-MASC-DU-verb-FEM-PL-IMP-No & {1.} \\
noun-FEM-PL-verb-FEM-PL-IMP-Yes & {1.} \\
\bottomrule
\end{tabular}}
    \caption{GF2 Accuracies of a few evaluation unit for \textbf{ara-com-2}.}
    \label{tab:eu_example}
\end{table}

\blue{
\section{Template-Aggregated Tables}
\label{sec:appendixE}
Table~\ref{tab:temp_agg} presents the scores for both \texttt{Generation} and \texttt{Judgement} scenarios across all LLMs for each template.
Table~\ref{tab:diff_temp_agg} shows the score differences between Direct and CoT modes for each LLM.
Table~\ref{tab:think_diff_temp_agg} displays the score differences between "Thinking" and "Non-Thinking" modes for Qwen3 across all templates.
\begin{table*}
    \centering
    \resizebox{\textwidth}{!}{
    \begin{tabular}{l|ccc||ccc||ccc||ccc||ccc||ccc||ccc||ccc}
    \toprule
  & \multicolumn{3}{c}{GFL2} & \multicolumn{3}{c}{GF2} & \multicolumn{3}{c}{Aya} & \multicolumn{3}{c}{Phi4} & \multicolumn{3}{c}{EuroLLM} & \multicolumn{3}{c}{Gem3} & \multicolumn{3}{c}{Qwen2.5} & \multicolumn{3}{c}{Qwen3} \\

 & Gen & JY & JN & Gen & JY & JN & Gen & JY & JN & Gen & JY & JN & Gen & JY & JN & Gen & JY & JN & Gen & JY & JN & Gen & JY & JN \\
\midrule
eng-com-1 & \gblue{1.} & \gblue{1.} & \gblue{1.} & \gblue{.985} & \gblue{1.} & \gblue{1.} & \gblue{.889} & \gblue{.967} & \gblue{.916} & \gblue{1.} & \gblue{.984} & \gblue{.913} & \gblue{1.} & \gblue{.979} & \gblue{.519} & \gblue{.995} & \gblue{1.} & \gblue{1.} & \gblue{1.} & \gblue{.990} & \gblue{1.} & \gblue{1.} & \gblue{1.} & \gblue{1.} \\
eng-com-2 & \gblue{1.} & \gblue{1.} & N/A & \gblue{1.} & \gblue{1.} & N/A & \gblue{.880} & \gblue{.731} & N/A & \gblue{1.} & \gblue{.990} & N/A & \gblue{1.} & \gblue{.980} & N/A & \gblue{1.} & \gblue{.970} & N/A & \gblue{1.} & \gblue{.947} & N/A & \gblue{1.} & \gblue{1.} & N/A \\
eng-com-3 & \gblue{1.} & \gblue{.969} & N/A & \gblue{1.} & \gblue{.990} & N/A & \gblue{.990} & \gblue{.807} & N/A & \gblue{.990} & \gblue{.762} & N/A & \gblue{.880} & \gblue{.900} & N/A & \gblue{1.} & \gblue{.960} & N/A & \gblue{.990} & \gblue{.881} & N/A & \gblue{1.} & \gblue{.980} & N/A \\
\midrule
ara-com-1 & \gblue{.975} & \gblue{.944} & \gblue{0} & \gblue{.893} & \gblue{.972} & \gblue{.915} & \gblue{.360} & \gblue{.925} & \gblue{.821} & \gblue{0} & \gblue{.842} & \gblue{.192} & \gblue{.175} & \gblue{.975} & \gblue{0} & \gblue{.408} & \gblue{.881} & \gblue{.740} & \gblue{.602} & \gblue{.838} & \gblue{.784} & \gblue{.625} & \gblue{.830} & \gblue{.705} \\
ara-com-2 & \gblue{.186} & \gblue{.973} & \gblue{.235} & \gblue{.928} & \gblue{.986} & \gblue{.837} & \gblue{.397} & \gblue{.667} & \gblue{.776} & \gblue{0} & \gblue{.824} & \gblue{.136} & \gblue{0} & \gblue{.990} & \gblue{0} & \gblue{.412} & \gblue{.885} & \gblue{.459} & \gblue{.160} & \gblue{.355} & \gblue{.867} & \gblue{.303} & \gblue{.753} & \gblue{.549} \\
ara-com-3 & \gblue{.848} & \gblue{.915} & \gblue{.915} & \gblue{.951} & \gblue{.889} & \gblue{.887} & \gblue{.559} & \gblue{.886} & \gblue{.708} & \gblue{.192} & \gblue{.865} & \gblue{.241} & \gblue{.296} & \gblue{.897} & \gblue{.136} & \gblue{.168} & \gblue{.489} & \gblue{.847} & \gblue{.437} & \gblue{.659} & \gblue{.888} & \gblue{0} & \gblue{.576} & \gblue{.802} \\
ara-1 & \gblue{.787} & \gblue{.541} & \gblue{.470} & \gblue{.500} & \gblue{.555} & \gblue{.343} & \gblue{.713} & \gblue{.800} & \gblue{.533} & \gblue{.305} & \gblue{.820} & \gblue{.107} & \gblue{.417} & \gblue{.850} & \gblue{.152} & \gblue{.864} & \gblue{.439} & \gblue{.492} & \gblue{.591} & \gblue{.836} & \gblue{.267} & \gblue{.649} & \gblue{.713} & \gblue{.738} \\
\midrule
heb-com-1 & \gblue{.956} & \gblue{.896} & \gblue{.920} & \gblue{.961} & \gblue{.935} & \gblue{.954} & \gblue{.606} & \gblue{.929} & \gblue{.891} & \gblue{0} & \gblue{.893} & \gblue{.306} & \gblue{0} & \gblue{.878} & \gblue{.065} & \gblue{.882} & \gblue{.931} & \gblue{.904} & \gblue{.523} & \gblue{.810} & \gblue{.929} & \gblue{.612} & \gblue{.841} & \gblue{.807} \\
heb-com-2 & \gblue{.858} & \gblue{.824} & \gblue{.714} & \gblue{.896} & \gblue{.761} & \gblue{.891} & \gblue{.345} & \gblue{.319} & \gblue{.942} & \gblue{0} & \gblue{.975} & \gblue{.036} & \gblue{.160} & \gblue{.955} & \gblue{0} & \gblue{.718} & \gblue{.820} & \gblue{.813} & \gblue{.546} & \gblue{.366} & \gblue{.872} & \gblue{.518} & \gblue{.770} & \gblue{.656} \\
heb-com-3 & \gblue{.974} & \gblue{.987} & \gblue{.936} & \gblue{.990} & \gblue{.992} & \gblue{.959} & \gblue{.919} & \gblue{.874} & \gblue{.947} & \gblue{.717} & \gblue{.922} & \gblue{.319} & \gblue{.611} & \gblue{.835} & \gblue{.208} & \gblue{.950} & \gblue{.909} & \gblue{.920} & \gblue{.788} & \gblue{.792} & \gblue{.924} & \gblue{.749} & \gblue{.924} & \gblue{.922} \\
heb-1 & \gblue{.726} & \gblue{.969} & \gblue{.795} & \gblue{.669} & \gblue{.723} & \gblue{.617} & \gblue{.719} & \gblue{.949} & \gblue{.611} & \gblue{.319} & \gblue{.947} & \gblue{.048} & \gblue{.141} & \gblue{.916} & \gblue{.137} & \gblue{.566} & \gblue{.940} & \gblue{.467} & \gblue{.401} & \gblue{.840} & \gblue{.352} & \gblue{.457} & \gblue{.838} & \gblue{.684} \\
\midrule
rus-com-1 & \gblue{.954} & \gblue{1.} & \gblue{.999} & \gblue{.990} & \gblue{1.} & \gblue{.999} & \gblue{.851} & \gblue{.990} & \gblue{.948} & \gblue{.747} & \gblue{.997} & \gblue{.722} & \gblue{.801} & \gblue{.985} & \gblue{.119} & \gblue{.899} & \gblue{.987} & \gblue{.995} & \gblue{.797} & \gblue{.997} & \gblue{.986} & \gblue{.898} & \gblue{.982} & \gblue{.986} \\
rus-com-2 & \gblue{.990} & \gblue{.990} & \gblue{.472} & \gblue{.940} & \gblue{.990} & \gblue{.778} & \gblue{.818} & \gblue{.950} & \gblue{.582} & \gblue{.618} & \gblue{.979} & \gblue{0} & \gblue{.854} & \gblue{1.} & \gblue{0} & \gblue{.699} & \gblue{.990} & \gblue{.894} & \gblue{.676} & \gblue{.950} & \gblue{.879} & \gblue{.827} & \gblue{.980} & \gblue{.875} \\
rus-com-3 & \gblue{.880} & \gblue{.907} & \gblue{0} & \gblue{.967} & \gblue{.973} & \gblue{0} & \gblue{.356} & \gblue{.898} & \gblue{.786} & \gblue{0} & \gblue{.920} & \gblue{0} & \gblue{0} & \gblue{.972} & \gblue{0} & \gblue{.902} & \gblue{.508} & \gblue{.942} & \gblue{.812} & \gblue{.473} & \gblue{.979} & \gblue{.694} & \gblue{.942} & \gblue{.927} \\
rus-1 & \gblue{.076} & \gblue{.734} & \gblue{.198} & \gblue{.789} & \gblue{.711} & \gblue{.246} & \gblue{.458} & \gblue{.789} & \gblue{.192} & \gblue{.207} & \gblue{.940} & \gblue{0} & \gblue{.429} & \gblue{.990} & \gblue{.034} & \gblue{.303} & \gblue{.546} & \gblue{.355} & \gblue{.323} & \gblue{.587} & \gblue{.335} & \gblue{.359} & \gblue{.597} & \gblue{.420} \\
\midrule
tur-com-1 & \gblue{.359} & \gblue{1.} & \gblue{.720} & \gblue{.964} & \gblue{.990} & \gblue{.683} & \gblue{.183} & \gblue{.953} & \gblue{.436} & \gblue{.401} & \gblue{.973} & \gblue{.246} & \gblue{.484} & \gblue{.949} & \gblue{0} & \gblue{.466} & \gblue{.965} & \gblue{.545} & \gblue{.576} & \gblue{.984} & \gblue{.770} & \gblue{.446} & \gblue{.970} & \gblue{.798} \\
tur-com-2 & \gblue{1.} & \gblue{1.} & \gblue{.755} & \gblue{.925} & \gblue{.990} & \gblue{.898} & \gblue{.847} & \gblue{.909} & \gblue{.670} & \gblue{.073} & \gblue{.969} & \gblue{.155} & \gblue{.873} & \gblue{.990} & \gblue{0} & \gblue{.889} & \gblue{.990} & \gblue{.468} & \gblue{.624} & \gblue{.881} & \gblue{.302} & \gblue{.723} & \gblue{.980} & \gblue{.587} \\
tur-com-3 & \gblue{.954} & \gblue{.979} & \gblue{.407} & \gblue{1.} & \gblue{.985} & \gblue{.882} & \gblue{.863} & \gblue{.698} & \gblue{.827} & \gblue{.934} & \gblue{.902} & \gblue{.449} & \gblue{.763} & \gblue{.889} & \gblue{.294} & \gblue{.443} & \gblue{.914} & \gblue{.843} & \gblue{.790} & \gblue{.949} & \gblue{.867} & \gblue{.611} & \gblue{.959} & \gblue{.857} \\
tur-1 & \gblue{.824} & \gblue{1.} & \gblue{.118} & \gblue{1.} & \gblue{1.} & \gblue{.471} & \gblue{.941} & \gblue{.882} & \gblue{.529} & \gblue{.529} & \gblue{.765} & \gblue{.294} & \gblue{.353} & \gblue{1.} & \gblue{.294} & \gblue{.647} & \gblue{1.} & \gblue{.294} & \gblue{.471} & \gblue{.882} & \gblue{.529} & \gblue{.824} & \gblue{1.} & \gblue{.471} \\
tur-2 & \gblue{.949} & \gblue{1.} & \gblue{.698} & \gblue{.940} & \gblue{.936} & \gblue{.643} & \gblue{.035} & \gblue{.659} & \gblue{.586} & \gblue{.123} & \gblue{.969} & \gblue{0} & \gblue{0} & \gblue{1.} & \gblue{0} & \gblue{.416} & \gblue{.970} & \gblue{.446} & \gblue{.057} & \gblue{.936} & \gblue{0} & \gblue{.178} & \gblue{.738} & \gblue{.635} \\

\midrule
fin-com-1 & \gblue{1.} & \gblue{1.} & \gblue{.990} & \gblue{1.} & \gblue{1.} & \gblue{.980} & \gblue{.558} & \gblue{.940} & \gblue{.899} & \gblue{.688} & \gblue{.945} & \gblue{.672} & \gblue{.788} & \gblue{1.} & \gblue{.149} & \gblue{.954} & \gblue{1.} & \gblue{.985} & \gblue{.687} & \gblue{.985} & \gblue{.968} & \gblue{.772} & \gblue{.990} & \gblue{.956} \\
fin-com-2 & \gblue{.969} & \gblue{1.} & \gblue{.760} & \gblue{.950} & \gblue{1.} & \gblue{.889} & \gblue{.509} & \gblue{.836} & \gblue{.463} & \gblue{.164} & \gblue{.936} & \gblue{.194} & \gblue{.836} & \gblue{1.} & \gblue{.039} & \gblue{.838} & \gblue{1.} & \gblue{.720} & \gblue{.327} & \gblue{.990} & \gblue{.181} & \gblue{.549} & \gblue{1.} & \gblue{.750} \\
fin-com-3 & \gblue{.581} & \gblue{.943} & \gblue{.828} & \gblue{.919} & \gblue{.980} & \gblue{.941} & \gblue{0} & \gblue{.683} & \gblue{.662} & \gblue{.059} & \gblue{.665} & \gblue{.351} & \gblue{.142} & \gblue{.951} & \gblue{0} & \gblue{.188} & \gblue{.757} & \gblue{.930} & \gblue{.343} & \gblue{.733} & \gblue{.932} & \gblue{.399} & \gblue{.851} & \gblue{.833} \\
fin-1 & \gblue{0} & \gblue{.938} & \gblue{0} & \gblue{.105} & \gblue{1.} & \gblue{0} & \gblue{0} & \gblue{.828} & \gblue{0} & \gblue{0} & \gblue{1.} & \gblue{0} & \gblue{.585} & \gblue{1.} & \gblue{0} & \gblue{0} & \gblue{.970} & \gblue{0} & \gblue{.485} & \gblue{.741} & \gblue{.261} & \gblue{0} & \gblue{.622} & \gblue{.320} \\
fin-2 & \gblue{.022} & \gblue{.956} & \gblue{.689} & \gblue{.911} & \gblue{.978} & \gblue{.489} & \gblue{0} & \gblue{.200} & \gblue{.578} & \gblue{.022} & \gblue{.733} & \gblue{.533} & \gblue{.200} & \gblue{1.} & \gblue{0} & \gblue{0} & \gblue{.911} & \gblue{.378} & \gblue{0} & \gblue{.889} & \gblue{.356} & \gblue{.178} & \gblue{.689} & \gblue{.533} \\
\bottomrule
\end{tabular}
}
\caption{The scores for Generation (Gen) and Judgement for positive (JY) and negative (JN) instances for each LLM on each template. The darker the color (the higher the score), the better.}
\label{tab:temp_agg}
\end{table*}
\begin{table*}
    \centering
    \resizebox{\textwidth}{!}{
    \begin{tabular}{l|ccc||ccc||ccc||ccc||ccc||ccc||ccc||ccc}
\toprule
  & \multicolumn{3}{c}{GFL2} & \multicolumn{3}{c}{GF2} & \multicolumn{3}{c}{Aya} & \multicolumn{3}{c}{Phi4} & \multicolumn{3}{c}{EuroLLM} & \multicolumn{3}{c}{Gem3} & \multicolumn{3}{c}{Qwen2.5} & \multicolumn{3}{c}{Qwen3} \\

 & Gen & JY & JN & Gen & JY & JN & Gen & JY & JN & Gen & JY & JN & Gen & JY & JN & Gen & JY & JN & Gen & JY & JN & Gen & JY & JN \\
\midrule
eng-com-1 & \gdelta{0} & \gdelta{0} & \gdelta{0} & \gdelta{-.015} & \gdelta{0} & \gdelta{0} & \gdelta{-.106} & \gdelta{-.033} & \gdelta{-.084} & \gdelta{.015} & \gdelta{.437} & \gdelta{-.087} & \gdelta{.036} & \gdelta{-.021} & \gdelta{-.461} & \gdelta{-.005} & \gdelta{0} & \gdelta{.010} & \gdelta{0} & \gdelta{.084} & \gdelta{0} & \gdelta{.020} & \gdelta{.016} & \gdelta{0} \\
eng-com-2 & \gdelta{0} & \gdelta{.120} & N/A & \gdelta{0} & \gdelta{.020} & N/A & \gdelta{-.120} & \gdelta{.439} & N/A & \gdelta{0} & \gdelta{.546} & N/A & \gdelta{0} & \gdelta{-.020} & N/A & \gdelta{0} & \gdelta{.083} & N/A & \gdelta{0} & \gdelta{.257} & N/A & \gdelta{0} & \gdelta{.095} & N/A \\
eng-com-3 & \gdelta{0} & \gdelta{.485} & N/A & \gdelta{.010} & \gdelta{.041} & N/A & \gdelta{.142} & \gdelta{.162} & N/A & \gdelta{-.010} & \gdelta{.117} & N/A & \gdelta{-.120} & \gdelta{-.100} & N/A & \gdelta{0} & \gdelta{.350} & N/A & \gdelta{.054} & \gdelta{.881} & N/A & \gdelta{0} & \gdelta{.467} & N/A \\
\midrule
ara-com-1 & \gdelta{.677} & \gdelta{-.001} & \gdelta{0} & \gdelta{.893} & \gdelta{.270} & \gdelta{.114} & \gdelta{.360} & \gdelta{.222} & \gdelta{.002} & \gdelta{0} & \gdelta{.389} & \gdelta{-.385} & \gdelta{.175} & \gdelta{.033} & \gdelta{0} & \gdelta{.408} & \gdelta{-.014} & \gdelta{.356} & \gdelta{.602} & \gdelta{.385} & \gdelta{-.158} & \gdelta{.625} & \gdelta{.320} & \gdelta{-.016} \\
ara-com-2 & \gdelta{-.702} & \gdelta{-.002} & \gdelta{.235} & \gdelta{.005} & \gdelta{.097} & \gdelta{.581} & \gdelta{.397} & \gdelta{.305} & \gdelta{-.105} & \gdelta{0} & \gdelta{-.105} & \gdelta{.136} & \gdelta{0} & \gdelta{.007} & \gdelta{0} & \gdelta{-.317} & \gdelta{-.036} & \gdelta{.344} & \gdelta{.160} & \gdelta{-.220} & \gdelta{.183} & \gdelta{.303} & \gdelta{.115} & \gdelta{.197} \\
ara-com-3 & \gdelta{-.056} & \gdelta{.033} & \gdelta{.321} & \gdelta{.015} & \gdelta{.113} & \gdelta{.045} & \gdelta{-.189} & \gdelta{.105} & \gdelta{-.081} & \gdelta{-.068} & \gdelta{.115} & \gdelta{-.144} & \gdelta{-.037} & \gdelta{-.044} & \gdelta{.136} & \gdelta{-.683} & \gdelta{-.406} & \gdelta{.475} & \gdelta{-.382} & \gdelta{.049} & \gdelta{-.007} & \gdelta{-.532} & \gdelta{-.091} & \gdelta{-.034} \\
ara-1 & \gdelta{-.182} & \gdelta{-.123} & \gdelta{.395} & \gdelta{-.449} & \gdelta{.063} & \gdelta{-.254} & \gdelta{.006} & \gdelta{.025} & \gdelta{.365} & \gdelta{-.208} & \gdelta{.061} & \gdelta{.010} & \gdelta{.175} & \gdelta{-.010} & \gdelta{.038} & \gdelta{-.076} & \gdelta{-.350} & \gdelta{.252} & \gdelta{-.136} & \gdelta{.259} & \gdelta{-.051} & \gdelta{-.090} & \gdelta{.075} & \gdelta{.318} \\
\midrule
heb-com-1 & \gdelta{-.005} & \gdelta{-.070} & \gdelta{.275} & \gdelta{.078} & \gdelta{.077} & \gdelta{.077} & \gdelta{-.208} & \gdelta{.006} & \gdelta{.240} & \gdelta{-.118} & \gdelta{.100} & \gdelta{-.179} & \gdelta{0} & \gdelta{-.122} & \gdelta{.065} & \gdelta{.055} & \gdelta{-.023} & \gdelta{.472} & \gdelta{.108} & \gdelta{.453} & \gdelta{-.039} & \gdelta{.315} & \gdelta{.209} & \gdelta{-.091} \\
heb-com-2 & \gdelta{.089} & \gdelta{.208} & \gdelta{.084} & \gdelta{-.027} & \gdelta{.337} & \gdelta{.039} & \gdelta{.345} & \gdelta{.049} & \gdelta{-.021} & \gdelta{0} & \gdelta{.079} & \gdelta{.036} & \gdelta{.160} & \gdelta{-.045} & \gdelta{0} & \gdelta{-.116} & \gdelta{.071} & \gdelta{.092} & \gdelta{.338} & \gdelta{.051} & \gdelta{.007} & \gdelta{-.030} & \gdelta{.325} & \gdelta{-.155} \\
heb-com-3 & \gdelta{-.016} & \gdelta{.093} & \gdelta{.061} & \gdelta{-.005} & \gdelta{.151} & \gdelta{-.018} & \gdelta{.035} & \gdelta{.094} & \gdelta{-.028} & \gdelta{.082} & \gdelta{-.001} & \gdelta{-.137} & \gdelta{.241} & \gdelta{-.165} & \gdelta{.208} & \gdelta{-.022} & \gdelta{.021} & \gdelta{-.042} & \gdelta{.121} & \gdelta{.276} & \gdelta{-.057} & \gdelta{.118} & \gdelta{.185} & \gdelta{-.007} \\
heb-1 & \gdelta{-.254} & \gdelta{.189} & \gdelta{.305} & \gdelta{-.311} & \gdelta{-.097} & \gdelta{.057} & \gdelta{.195} & \gdelta{-.041} & \gdelta{.453} & \gdelta{-.020} & \gdelta{.009} & \gdelta{0} & \gdelta{-.135} & \gdelta{-.084} & \gdelta{.137} & \gdelta{-.178} & \gdelta{.089} & \gdelta{.241} & \gdelta{-.107} & \gdelta{.030} & \gdelta{.117} & \gdelta{-.054} & \gdelta{-.015} & \gdelta{.378} \\
\midrule
rus-com-1 & \gdelta{.032} & \gdelta{.003} & \gdelta{0} & \gdelta{.020} & \gdelta{0} & \gdelta{.002} & \gdelta{-.026} & \gdelta{.028} & \gdelta{-.052} & \gdelta{-.072} & \gdelta{.133} & \gdelta{-.228} & \gdelta{.006} & \gdelta{-.015} & \gdelta{-.675} & \gdelta{-.034} & \gdelta{.011} & \gdelta{0} & \gdelta{-.100} & \gdelta{.221} & \gdelta{-.014} & \gdelta{.028} & \gdelta{.107} & \gdelta{-.014} \\
rus-com-2 & \gdelta{.096} & \gdelta{.030} & \gdelta{-.107} & \gdelta{.117} & \gdelta{-.010} & \gdelta{.619} & \gdelta{-.036} & \gdelta{-.010} & \gdelta{.236} & \gdelta{-.160} & \gdelta{-.001} & \gdelta{0} & \gdelta{-.004} & \gdelta{0} & \gdelta{0} & \gdelta{-.197} & \gdelta{.030} & \gdelta{.390} & \gdelta{-.170} & \gdelta{.146} & \gdelta{.018} & \gdelta{.035} & \gdelta{.122} & \gdelta{.035} \\
rus-com-3 & \gdelta{-.081} & \gdelta{.211} & \gdelta{-.997} & \gdelta{-.014} & \gdelta{.150} & \gdelta{-.992} & \gdelta{.356} & \gdelta{.630} & \gdelta{-.207} & \gdelta{0} & \gdelta{.190} & \gdelta{0} & \gdelta{-.180} & \gdelta{-.024} & \gdelta{0} & \gdelta{-.017} & \gdelta{.048} & \gdelta{-.032} & \gdelta{.113} & \gdelta{.473} & \gdelta{-.018} & \gdelta{.520} & \gdelta{.442} & \gdelta{-.036} \\
rus-1 & \gdelta{-.351} & \gdelta{-.042} & \gdelta{.091} & \gdelta{-.075} & \gdelta{.049} & \gdelta{.133} & \gdelta{-.402} & \gdelta{.015} & \gdelta{.057} & \gdelta{.207} & \gdelta{.100} & \gdelta{-.198} & \gdelta{-.009} & \gdelta{-.010} & \gdelta{.034} & \gdelta{.089} & \gdelta{.080} & \gdelta{-.088} & \gdelta{.072} & \gdelta{.121} & \gdelta{-.004} & \gdelta{.217} & \gdelta{.046} & \gdelta{.015} \\
\midrule
tur-com-1 & \gdelta{-.526} & \gdelta{.031} & \gdelta{.601} & \gdelta{.027} & \gdelta{.037} & \gdelta{.395} & \gdelta{-.324} & \gdelta{.095} & \gdelta{.303} & \gdelta{-.050} & \gdelta{.036} & \gdelta{.166} & \gdelta{.119} & \gdelta{-.041} & \gdelta{0} & \gdelta{-.385} & \gdelta{-.020} & \gdelta{.545} & \gdelta{-.092} & \gdelta{.401} & \gdelta{-.036} & \gdelta{-.165} & \gdelta{.051} & \gdelta{.313} \\
tur-com-2 & \gdelta{.010} & \gdelta{.031} & \gdelta{.308} & \gdelta{-.055} & \gdelta{-.010} & \gdelta{.418} & \gdelta{.023} & \gdelta{.004} & \gdelta{.438} & \gdelta{.034} & \gdelta{0} & \gdelta{.155} & \gdelta{.033} & \gdelta{-.010} & \gdelta{0} & \gdelta{-.111} & \gdelta{.043} & \gdelta{.430} & \gdelta{-.191} & \gdelta{.282} & \gdelta{-.376} & \gdelta{-.052} & \gdelta{.143} & \gdelta{-.055} \\
tur-com-3 & \gdelta{.005} & \gdelta{.095} & \gdelta{-.264} & \gdelta{.020} & \gdelta{.056} & \gdelta{.009} & \gdelta{-.101} & \gdelta{-.071} & \gdelta{-.158} & \gdelta{-.025} & \gdelta{-.027} & \gdelta{-.192} & \gdelta{-.175} & \gdelta{-.106} & \gdelta{.139} & \gdelta{-.531} & \gdelta{.076} & \gdelta{-.053} & \gdelta{-.129} & \gdelta{.520} & \gdelta{-.128} & \gdelta{-.379} & \gdelta{.238} & \gdelta{-.097} \\
tur-1 & \gdelta{0} & \gdelta{0} & \gdelta{-.058} & \gdelta{.176} & \gdelta{0} & \gdelta{.177} & \gdelta{.588} & \gdelta{.058} & \gdelta{-.059} & \gdelta{.176} & \gdelta{-.176} & \gdelta{.294} & \gdelta{.177} & \gdelta{0} & \gdelta{.294} & \gdelta{-.118} & \gdelta{0} & \gdelta{.118} & \gdelta{.236} & \gdelta{.117} & \gdelta{.058} & \gdelta{0} & \gdelta{0} & \gdelta{.118} \\
tur-2 & \gdelta{.382} & \gdelta{.040} & \gdelta{.391} & \gdelta{.090} & \gdelta{-.054} & \gdelta{.007} & \gdelta{-.102} & \gdelta{.606} & \gdelta{-.400} & \gdelta{.123} & \gdelta{.763} & \gdelta{-.698} & \gdelta{-.033} & \gdelta{.767} & \gdelta{-.747} & \gdelta{-.078} & \gdelta{.734} & \gdelta{-.355} & \gdelta{-.005} & \gdelta{.936} & \gdelta{-.990} & \gdelta{.058} & \gdelta{.175} & \gdelta{-.097} \\
\midrule
fin-com-1 & \gdelta{.036} & \gdelta{.010} & \gdelta{-.010} & \gdelta{.005} & \gdelta{.005} & \gdelta{.007} & \gdelta{-.136} & \gdelta{.167} & \gdelta{.274} & \gdelta{.030} & \gdelta{.035} & \gdelta{-.079} & \gdelta{-.136} & \gdelta{0} & \gdelta{.149} & \gdelta{.007} & \gdelta{0} & \gdelta{-.005} & \gdelta{.011} & \gdelta{.477} & \gdelta{-.032} & \gdelta{0} & \gdelta{.048} & \gdelta{.001} \\
fin-com-2 & \gdelta{-.011} & \gdelta{0} & \gdelta{.322} & \gdelta{-.019} & \gdelta{0} & \gdelta{.556} & \gdelta{.168} & \gdelta{-.154} & \gdelta{.463} & \gdelta{.020} & \gdelta{-.044} & \gdelta{.002} & \gdelta{.379} & \gdelta{0} & \gdelta{.039} & \gdelta{-.087} & \gdelta{0} & \gdelta{.607} & \gdelta{.183} & \gdelta{.130} & \gdelta{-.357} & \gdelta{.150} & \gdelta{.031} & \gdelta{.445} \\
fin-com-3 & \gdelta{.322} & \gdelta{.073} & \gdelta{-.135} & \gdelta{.008} & \gdelta{.023} & \gdelta{.365} & \gdelta{0} & \gdelta{-.031} & \gdelta{.383} & \gdelta{.059} & \gdelta{-.027} & \gdelta{-.260} & \gdelta{-.171} & \gdelta{-.047} & \gdelta{0} & \gdelta{-.513} & \gdelta{-.016} & \gdelta{-.041} & \gdelta{.216} & \gdelta{.733} & \gdelta{-.043} & \gdelta{.399} & \gdelta{-.036} & \gdelta{.013} \\
fin-1 & \gdelta{-1.} & \gdelta{.557} & \gdelta{-.867} & \gdelta{.105} & \gdelta{0} & \gdelta{0} & \gdelta{-.564} & \gdelta{-.142} & \gdelta{0} & \gdelta{0} & \gdelta{0} & \gdelta{0} & \gdelta{-.318} & \gdelta{0} & \gdelta{0} & \gdelta{-.970} & \gdelta{0} & \gdelta{0} & \gdelta{.115} & \gdelta{.119} & \gdelta{.156} & \gdelta{0} & \gdelta{.432} & \gdelta{-.618} \\
fin-2 & \gdelta{-.956} & \gdelta{0} & \gdelta{-.222} & \gdelta{.133} & \gdelta{.045} & \gdelta{-.400} & \gdelta{-.333} & \gdelta{-.667} & \gdelta{.511} & \gdelta{.022} & \gdelta{.066} & \gdelta{-.045} & \gdelta{-.511} & \gdelta{0} & \gdelta{0} & \gdelta{-.911} & \gdelta{-.045} & \gdelta{.156} & \gdelta{-.644} & \gdelta{-.022} & \gdelta{-.044} & \gdelta{-.489} & \gdelta{.067} & \gdelta{-.245} \\

\bottomrule
\end{tabular}

}

\caption{Score differences for Generation (Gen), and Judgements of positive (JY) and negative (JN) instances for each LLM when shifting from Direct prompting to Chain-of-Thought (CoT).}
\label{tab:diff_temp_agg}
\end{table*}
\begin{table}[t]
\small
    \centering
\begin{tabular}{lrrr}
\toprule
 & \multicolumn{3}{c}{Qwen3} \\
 & Gen & JY & JN \\
\midrule
eng-com-1 & \gdelta{.020} & \gdelta{.011} & \gdelta{0} \\
eng-com-2 & \gdelta{0} & \gdelta{.095} & N/A \\
eng-com-3 & \gdelta{0} & \gdelta{.434} & N/A \\
\midrule
ara-com-1 & \gdelta{.619} & \gdelta{.252} & \gdelta{.164} \\
ara-com-2 & \gdelta{.366} & \gdelta{-.016} & \gdelta{.416} \\
ara-com-3 & \gdelta{-.385} & \gdelta{-.143} & \gdelta{.046} \\
ara-1 & \gdelta{-.064} & \gdelta{-.052} & \gdelta{.149} \\
\midrule
heb-com-1 & \gdelta{.352} & \gdelta{.203} & \gdelta{-.115} \\
heb-com-2 & \gdelta{.008} & \gdelta{.281} & \gdelta{-.015} \\
heb-com-3 & \gdelta{.120} & \gdelta{.144} & \gdelta{.023} \\
heb-1 & \gdelta{.088} & \gdelta{-.053} & \gdelta{.195} \\
\midrule
rus-com-1 & \gdelta{.013} & \gdelta{.100} & \gdelta{0} \\
rus-com-2 & \gdelta{.097} & \gdelta{.086} & \gdelta{.125} \\
rus-com-3 & \gdelta{.322} & \gdelta{.379} & \gdelta{.015} \\
rus-1 & \gdelta{.173} & \gdelta{.085} & \gdelta{-.111} \\
\midrule
tur-com-1 & \gdelta{-.007} & \gdelta{-.027} & \gdelta{.418} \\
tur-com-2 & \gdelta{-.046} & \gdelta{.153} & \gdelta{-.055} \\
tur-com-3 & \gdelta{-.212} & \gdelta{.198} & \gdelta{.011} \\
tur-1 & \gdelta{-.118} & \gdelta{-.118} & \gdelta{.529} \\
tur-2 & \gdelta{.285} & \gdelta{.189} & \gdelta{.017} \\

\midrule
fin-com-1 & \gdelta{-.076} & \gdelta{-.013} & \gdelta{-.012} \\
fin-com-2 & \gdelta{.134} & \gdelta{-.029} & \gdelta{.477} \\
fin-com-3 & \gdelta{.276} & \gdelta{-.034} & \gdelta{.103} \\
fin-1 & \gdelta{0} & \gdelta{.448} & \gdelta{-.938} \\
fin-2 & \gdelta{-.289} & \gdelta{.200} & \gdelta{-.156} \\

\bottomrule
\end{tabular}
\caption{Score difference between "Thinking" and "Non-Thinking" mode for Qwen3 on the template level.
}
\label{tab:think_diff_temp_agg}
\end{table}

}

\blue{
\section{Language-Aggregated Tables}
\label{sec:appendixF}
We present additional results aggregated at the language level. Table~\ref{tab:lang_agg} summarizes the overall performance of all LLMs across different languages. Similarly, Tables~\ref{tab:diff_lang_agg} and~\ref{tab:think_diff_lang_agg} show the performance gains and losses from Direct$\rightarrow$CoT and Direct$\rightarrow$Thinking Mode, respectively, across the languages.

\begin{table*}[t]
    \centering
    \resizebox{\textwidth}{!}{
\begin{tabular}{lrrrrr||rrrrr||rrrrr||rrrrr||rrrrr||rrrrr||rrrrr||rrrrr}
\toprule
  & \multicolumn{5}{c}{GFL2} & \multicolumn{5}{c}{GF2} & \multicolumn{5}{c}{Aya} & \multicolumn{5}{c}{Phi4} & \multicolumn{5}{c}{EuroLLM} & \multicolumn{5}{c}{Gem3} & \multicolumn{5}{c}{Qwen2.5} & \multicolumn{5}{c}{Qwen3} \\
 & Gen & JY & JN & Jud & MS & Gen & JY & JN & Jud & MS & Gen & JY & JN & Jud & MS & Gen & JY & JN & Jud & MS & Gen & JY & JN & Jud & MS & Gen & JY & JN & Jud & MS & Gen & JY & JN & Jud & MS & Gen & JY & JN & Jud & MS \\
\midrule
eng & \gblue{1.} & \gblue{.990} & \gblue{1.} & \gblue{.995} & \gblue{.997} & \gblue{.995} & \gblue{.997} & \gblue{1.} & \gblue{.998} & \gblue{.997} & \gblue{.920} & \gblue{.835} & \gblue{.916} & \gblue{.874} & \gblue{.896} & \gblue{.997} & \gblue{.912} & \gblue{.913} & \gblue{.912} & \gblue{.953} & \gblue{.960} & \gblue{.953} & \gblue{.519} & \gblue{.672} & \gblue{.791} & \gblue{.998} & \gblue{.977} & \gblue{1.} & \gblue{.988} & \gblue{.993} & \gblue{.997} & \gblue{.939} & \gblue{1.} & \gblue{.969} & \gblue{.982} & \gblue{1.} & \gblue{.993} & \gblue{1.} & \gblue{.997} & \gblue{.998} \\
ara & \gblue{.699} & \gblue{.843} & \gblue{.405} & \gblue{.547} & \gblue{.614} & \gblue{.818} & \gblue{.850} & \gblue{.746} & \gblue{.795} & \gblue{.806} & \gblue{.507} & \gblue{.820} & \gblue{.710} & \gblue{.761} & \gblue{.609} & \gblue{.124} & \gblue{.838} & \gblue{.169} & \gblue{.281} & \gblue{.172} & \gblue{.222} & \gblue{.928} & \gblue{.072} & \gblue{.134} & \gblue{.167} & \gblue{.463} & \gblue{.674} & \gblue{.635} & \gblue{.653} & \gblue{.542} & \gblue{.448} & \gblue{.672} & \gblue{.702} & \gblue{.686} & \gblue{.542} & \gblue{.394} & \gblue{.718} & \gblue{.698} & \gblue{.708} & \gblue{.507} \\
heb & \gblue{.878} & \gblue{.919} & \gblue{.841} & \gblue{.878} & \gblue{.878} & \gblue{.879} & \gblue{.853} & \gblue{.855} & \gblue{.854} & \gblue{.866} & \gblue{.647} & \gblue{.768} & \gblue{.848} & \gblue{.806} & \gblue{.718} & \gblue{.259} & \gblue{.934} & \gblue{.177} & \gblue{.298} & \gblue{.277} & \gblue{.228} & \gblue{.896} & \gblue{.103} & \gblue{.184} & \gblue{.204} & \gblue{.779} & \gblue{.900} & \gblue{.776} & \gblue{.833} & \gblue{.805} & \gblue{.564} & \gblue{.702} & \gblue{.769} & \gblue{.734} & \gblue{.638} & \gblue{.584} & \gblue{.843} & \gblue{.767} & \gblue{.803} & \gblue{.676} \\
rus & \gblue{.725} & \gblue{.908} & \gblue{.417} & \gblue{.572} & \gblue{.639} & \gblue{.922} & \gblue{.918} & \gblue{.506} & \gblue{.652} & \gblue{.764} & \gblue{.621} & \gblue{.907} & \gblue{.627} & \gblue{.741} & \gblue{.676} & \gblue{.393} & \gblue{.959} & \gblue{.180} & \gblue{.304} & \gblue{.343} & \gblue{.521} & \gblue{.987} & \gblue{.038} & \gblue{.074} & \gblue{.129} & \gblue{.701} & \gblue{.758} & \gblue{.796} & \gblue{.777} & \gblue{.737} & \gblue{.652} & \gblue{.752} & \gblue{.795} & \gblue{.773} & \gblue{.707} & \gblue{.694} & \gblue{.875} & \gblue{.802} & \gblue{.837} & \gblue{.759} \\
tur & \gblue{.817} & \gblue{.996} & \gblue{.540} & \gblue{.700} & \gblue{.754} & \gblue{.966} & \gblue{.980} & \gblue{.715} & \gblue{.827} & \gblue{.891} & \gblue{.574} & \gblue{.820} & \gblue{.610} & \gblue{.699} & \gblue{.630} & \gblue{.412} & \gblue{.916} & \gblue{.229} & \gblue{.366} & \gblue{.388} & \gblue{.495} & \gblue{.966} & \gblue{.118} & \gblue{.210} & \gblue{.294} & \gblue{.572} & \gblue{.968} & \gblue{.519} & \gblue{.676} & \gblue{.620} & \gblue{.504} & \gblue{.926} & \gblue{.494} & \gblue{.644} & \gblue{.565} & \gblue{.556} & \gblue{.929} & \gblue{.670} & \gblue{.778} & \gblue{.649} \\
fin & \gblue{.514} & \gblue{.967} & \gblue{.653} & \gblue{.780} & \gblue{.620} & \gblue{.777} & \gblue{.992} & \gblue{.660} & \gblue{.792} & \gblue{.785} & \gblue{.213} & \gblue{.697} & \gblue{.520} & \gblue{.596} & \gblue{.314} & \gblue{.187} & \gblue{.856} & \gblue{.350} & \gblue{.497} & \gblue{.271} & \gblue{.510} & \gblue{.990} & \gblue{.038} & \gblue{.072} & \gblue{.127} & \gblue{.396} & \gblue{.928} & \gblue{.603} & \gblue{.731} & \gblue{.514} & \gblue{.368} & \gblue{.868} & \gblue{.540} & \gblue{.665} & \gblue{.474} & \gblue{.380} & \gblue{.830} & \gblue{.678} & \gblue{.747} & \gblue{.503} \\

\midrule
hmean & \gblue{.703} & \gblue{.924} & \gblue{.530} & \gblue{.673} & \gblue{.688} & \gblue{.867} & \gblue{.915} & \gblue{.676} & \gblue{.777} & \gblue{.820} & \gblue{.433} & \gblue{.796} & \gblue{.645} & \gblue{.713} & \gblue{.539} & \gblue{.225} & \gblue{.898} & \gblue{.206} & \gblue{.335} & \gblue{.269} & \gblue{.338} & \gblue{.952} & \gblue{.059} & \gblue{.111} & \gblue{.167} & \gblue{.547} & \gblue{.830} & \gblue{.649} & \gblue{.728} & \gblue{.625} & \gblue{.488} & \gblue{.772} & \gblue{.636} & \gblue{.697} & \gblue{.574} & \gblue{.494} & \gblue{.833} & \gblue{.719} & \gblue{.772} & \gblue{.603} \\

\bottomrule
\end{tabular}
}
\caption{The averaged scores over templates for Generation (Gen), Judgement for positive (JY) and negative (JN) instances for each LLM, aggregated Judgement (Jud), and aggregated model score (MS). The darker the color (the higher the score), the better.
}
\label{tab:lang_agg}
\end{table*}

\begin{table*}[t]
    \centering
    \resizebox{\textwidth}{!}{
\begin{tabular}{lrrrrr||rrrrr||rrrrr||rrrrr||rrrrr||rrrrr||rrrrr||rrrrr}
\toprule
  & \multicolumn{5}{c}{GFL2} & \multicolumn{5}{c}{GF2} & \multicolumn{5}{c}{Aya} & \multicolumn{5}{c}{Phi4} & \multicolumn{5}{c}{EuroLLM} & \multicolumn{5}{c}{Gem3} & \multicolumn{5}{c}{Qwen2.5} & \multicolumn{5}{c}{Qwen3} \\
 & Gen & JY & JN & Jud & MS & Gen & JY & JN & Jud & MS & Gen & JY & JN & Jud & MS & Gen & JY & JN & Jud & MS & Gen & JY & JN & Jud & MS & Gen & JY & JN & Jud & MS & Gen & JY & JN & Jud & MS & Gen & JY & JN & Jud & MS \\
\midrule
eng & \gdelta{0} & \gdelta{.202} & \gdelta{0} & \gdelta{.113} & \gdelta{.060} & \gdelta{-.002} & \gdelta{.020} & \gdelta{0} & \gdelta{.010} & \gdelta{.004} & \gdelta{-.028} & \gdelta{.189} & \gdelta{-.084} & \gdelta{.089} & \gdelta{.038} & \gdelta{.002} & \gdelta{.367} & \gdelta{-.087} & \gdelta{.207} & \gdelta{.127} & \gdelta{-.028} & \gdelta{-.047} & \gdelta{-.461} & \gdelta{-.318} & \gdelta{-.198} & \gdelta{-.002} & \gdelta{.144} & \gdelta{.010} & \gdelta{.084} & \gdelta{.043} & \gdelta{.018} & \gdelta{.407} & \gdelta{0} & \gdelta{.274} & \gdelta{.170} & \gdelta{.007} & \gdelta{.193} & \gdelta{0} & \gdelta{.107} & \gdelta{.060} \\
ara & \gdelta{-.066} & \gdelta{-.023} & \gdelta{.238} & \gdelta{.267} & \gdelta{.204} & \gdelta{.116} & \gdelta{.136} & \gdelta{.122} & \gdelta{.128} & \gdelta{.122} & \gdelta{.143} & \gdelta{.164} & \gdelta{.045} & \gdelta{.101} & \gdelta{.140} & \gdelta{-.069} & \gdelta{.115} & \gdelta{-.096} & \gdelta{-.106} & \gdelta{-.086} & \gdelta{.078} & \gdelta{-.004} & \gdelta{.044} & \gdelta{.078} & \gdelta{.087} & \gdelta{-.167} & \gdelta{-.202} & \gdelta{.357} & \gdelta{.232} & \gdelta{.037} & \gdelta{.061} & \gdelta{.118} & \gdelta{-.008} & \gdelta{.064} & \gdelta{.065} & \gdelta{.077} & \gdelta{.105} & \gdelta{.116} & \gdelta{.111} & \gdelta{.092} \\
heb & \gdelta{-.047} & \gdelta{.105} & \gdelta{.181} & \gdelta{.149} & \gdelta{.063} & \gdelta{-.066} & \gdelta{.117} & \gdelta{.039} & \gdelta{.080} & \gdelta{.015} & \gdelta{.092} & \gdelta{.027} & \gdelta{.161} & \gdelta{.093} & \gdelta{.093} & \gdelta{-.014} & \gdelta{.047} & \gdelta{-.070} & \gdelta{-.089} & \gdelta{-.043} & \gdelta{.066} & \gdelta{-.104} & \gdelta{.103} & \gdelta{.184} & \gdelta{.204} & \gdelta{-.065} & \gdelta{.039} & \gdelta{.191} & \gdelta{.137} & \gdelta{.042} & \gdelta{.115} & \gdelta{.202} & \gdelta{.007} & \gdelta{.131} & \gdelta{.123} & \gdelta{.087} & \gdelta{.176} & \gdelta{.031} & \gdelta{.104} & \gdelta{.095} \\
rus & \gdelta{-.076} & \gdelta{.051} & \gdelta{-.253} & \gdelta{-.181} & \gdelta{-.137} & \gdelta{.012} & \gdelta{.047} & \gdelta{-.060} & \gdelta{-.033} & \gdelta{-.018} & \gdelta{-.027} & \gdelta{.166} & \gdelta{.008} & \gdelta{.067} & \gdelta{.015} & \gdelta{-.006} & \gdelta{.106} & \gdelta{-.106} & \gdelta{-.126} & \gdelta{-.071} & \gdelta{-.047} & \gdelta{-.012} & \gdelta{-.160} & \gdelta{-.258} & \gdelta{-.289} & \gdelta{-.040} & \gdelta{.042} & \gdelta{.068} & \gdelta{.054} & \gdelta{.006} & \gdelta{-.021} & \gdelta{.240} & \gdelta{-.005} & \gdelta{.149} & \gdelta{.060} & \gdelta{.200} & \gdelta{.179} & \gdelta{0} & \gdelta{.092} & \gdelta{.165} \\
tur & \gdelta{-.026} & \gdelta{.039} & \gdelta{.196} & \gdelta{.194} & \gdelta{.122} & \gdelta{.052} & \gdelta{.006} & \gdelta{.201} & \gdelta{.154} & \gdelta{.116} & \gdelta{.017} & \gdelta{.138} & \gdelta{.025} & \gdelta{.070} & \gdelta{.039} & \gdelta{.052} & \gdelta{.119} & \gdelta{-.055} & \gdelta{-.052} & \gdelta{0} & \gdelta{.024} & \gdelta{.122} & \gdelta{-.063} & \gdelta{-.088} & \gdelta{-.070} & \gdelta{-.245} & \gdelta{.167} & \gdelta{.137} & \gdelta{.158} & \gdelta{-.014} & \gdelta{-.036} & \gdelta{.451} & \gdelta{-.294} & \gdelta{.051} & \gdelta{0} & \gdelta{-.108} & \gdelta{.121} & \gdelta{.036} & \gdelta{.068} & \gdelta{-.037} \\
fin & \gdelta{-.322} & \gdelta{.128} & \gdelta{-.182} & \gdelta{-.058} & \gdelta{-.217} & \gdelta{.046} & \gdelta{.015} & \gdelta{.106} & \gdelta{.085} & \gdelta{.066} & \gdelta{-.173} & \gdelta{-.165} & \gdelta{.326} & \gdelta{.279} & \gdelta{-.034} & \gdelta{.026} & \gdelta{.006} & \gdelta{-.076} & \gdelta{-.071} & \gdelta{.021} & \gdelta{-.151} & \gdelta{-.009} & \gdelta{.038} & \gdelta{.072} & \gdelta{.127} & \gdelta{-.495} & \gdelta{-.012} & \gdelta{.143} & \gdelta{.114} & \gdelta{-.215} & \gdelta{-.024} & \gdelta{.287} & \gdelta{-.064} & \gdelta{.074} & \gdelta{.003} & \gdelta{.012} & \gdelta{.108} & \gdelta{-.081} & \gdelta{.007} & \gdelta{.012} \\

\midrule
hmean & \gdelta{-.128} & \gdelta{.059} & \gdelta{.148} & \gdelta{.144} & \gdelta{.041} & \gdelta{.040} & \gdelta{.075} & \gdelta{.076} & \gdelta{.078} & \gdelta{.062} & \gdelta{-.045} & \gdelta{.067} & \gdelta{.208} & \gdelta{.166} & \gdelta{.029} & \gdelta{-.021} & \gdelta{.080} & \gdelta{-.086} & \gdelta{-.095} & \gdelta{-.044} & \gdelta{.068} & \gdelta{.001} & \gdelta{.059} & \gdelta{.111} & \gdelta{.167} & \gdelta{-.226} & \gdelta{-.002} & \gdelta{.213} & \gdelta{.156} & \gdelta{-.033} & \gdelta{.021} & \gdelta{.251} & \gdelta{-.089} & \gdelta{.091} & \gdelta{.047} & \gdelta{.056} & \gdelta{.138} & \gdelta{.027} & \gdelta{.078} & \gdelta{.065} \\

\bottomrule
\end{tabular}}
\caption{Score differences across languages for Generation (Gen), positive Judgement (JY), negative Judgement (JN), aggregated Judgement (Jud), and overall model score (MS) for each LLM when shifting from Direct prompting to CoT. Averages are computed using the arithmetic mean.}
\label{tab:diff_lang_agg}
\end{table*}

\begin{table}[H]
    \centering
\resizebox{\columnwidth}{!}{\begin{tabular}{lccccc}
\toprule
 & \multicolumn{5}{c}{Qwen3} \\
 & Gen & JY & JN & Jud & MS \\
\midrule
eng & \gdelta{.007} & \gdelta{.180} & \gdelta{0} & \gdelta{.101} & \gdelta{.057} \\
ara & \gdelta{.134} & \gdelta{.010} & \gdelta{.194} & \gdelta{.094} & \gdelta{.132} \\
heb & \gdelta{.142} & \gdelta{.144} & \gdelta{.022} & \gdelta{.084} & \gdelta{.123} \\
rus & \gdelta{.151} & \gdelta{.163} & \gdelta{.007} & \gdelta{.088} & \gdelta{.133} \\
tur & \gdelta{-.020} & \gdelta{.079} & \gdelta{.184} & \gdelta{.141} & \gdelta{.047} \\
fin & \gdelta{.009} & \gdelta{.114} & \gdelta{-.105} & \gdelta{-.006} & \gdelta{.007} \\

\midrule
hmean & \gdelta{.086} & \gdelta{.095} & \gdelta{.065} & \gdelta{.080} & \gdelta{.088} \\

\bottomrule
\end{tabular}}
\caption{Score differences between "Thinking" and "Non-Thinking" modes for Qwen3 on the language level.}
\label{tab:think_diff_lang_agg}
\end{table}

}


\blue{\section{IMPACT Checklist}
\label{sec:appendixK}
\begin{table*}[t]
    \centering
    \resizebox{\textwidth}{!}{
\begin{tabular}{l|lll||lll||lll||lll||lll||lll||lll||lll}
\toprule
 & \multicolumn{3}{c}{GFL2} & \multicolumn{3}{c}{GF2} & \multicolumn{3}{c}{Aya} & \multicolumn{3}{c}{Phi4} & \multicolumn{3}{c}{EuroLLM} & \multicolumn{3}{c}{Gem3} & \multicolumn{3}{c}{Qwen2.5} & \multicolumn{3}{c}{Qwen3} \\
 & Gen & JY & JN & Gen & JY & JN & Gen & JY & JN & Gen & JY & JN & Gen & JY & JN & Gen & JY & JN & Gen & JY & JN & Gen & JY & JN \\
 \toprule
eng-com-1 &  &  &  &  &  &  &  &  &  &  &  &  &  &  & X &  &  &  &  &  &  &  &  &  \\
eng-com-2 &  &  &  &  &  &  &  &  &  &  &  &  &  &  &  &  &  &  &  &  &  &  &  &  \\
eng-com-3 &  &  &  &  &  &  &  &  &  &  &  &  &  &  &  &  &  &  &  &  &  &  &  &  \\
\midrule
ara-com-1 &  &  & X &  &  & X & X &  & X & X &  & X & X &  & X & X &  & X & X &  & X & X &  & X \\
ara-com-2 & X &  & X &  &  & X & X & X & X & X &  & X & X &  & X & X & X & X & X & X & X & X & X & X \\
ara-com-3 & X & X & X &  & X & X & X &  & X & X &  & X & X &  & X & X & X & X & X & X & X & X & X & X \\
ara-1 &  & X & X & X & X & X &  &  & X & X &  & X & X &  & X &  & X & X & X &  & X &  &  &  \\
\midrule

heb-com-1 &  &  &  &  &  &  & X &  &  & X &  & X & X &  & X &  &  &  & X &  & X & X &  & X \\
heb-com-2 &  &  & X &  & X &  & X & X &  & X &  & X & X &  & X &  &  & X & X & X &  & X &  & X \\
heb-com-3 &  &  &  &  &  &  &  &  &  & X &  & X & X &  & X &  &  &  & X &  &  & X &  &  \\
heb-1 &  &  &  &  &  & X &  &  & X & X &  & X & X &  & X & X &  & X & X &  & X & X &  & X \\
\midrule

rus-com-1 &  &  &  &  &  &  &  &  &  &  &  & X &  &  & X &  &  &  &  &  &  &  &  &  \\
rus-com-2 &  &  & X &  &  &  &  &  & X & X &  & X &  &  & X &  &  &  & X &  &  &  &  &  \\
rus-com-3 & X &  & X &  &  & X & X &  & X & X & X & X & X &  & X & X & X & X & X & X & X & X &  & X \\
rus-1 & X &  & X &  & X & X & X &  & X & X &  & X & X &  & X & X & X & X & X & X & X & X & X & X \\
\midrule

tur-com-1 & X &  & X &  &  & X & X &  & X & X &  & X & X &  & X & X &  & X & X &  &  & X &  &  \\
tur-com-2 &  &  &  &  &  &  &  &  &  & X &  & X &  &  & X &  &  & X & X &  & X & X &  & X \\
tur-com-3 &  &  & X &  &  &  &  &  &  &  &  & X &  &  & X & X &  &  & X &  &  & X &  &  \\
tur-1 &  &  & X &  &  & X &  &  & X & X &  & X & X &  & X &  &  & X & X &  & X &  &  & X \\
tur-2 &  &  & X &  &  & X & X & X & X & X &  & X & X &  & X & X &  & X & X &  & X & X & X & X \\

\midrule

fin-com-1 &  &  &  &  &  &  & X &  &  & X &  & X &  &  & X &  &  &  &  &  &  &  &  &  \\
fin-com-2 &  &  &  &  &  &  & X &  & X & X &  & X &  &  & X &  &  &  & X &  & X & X &  &  \\
fin-com-3 & X &  & X &  &  & X & X & X & X & X & X & X & X &  & X & X & X &  & X & X & X & X &  & X \\
fin-1 & X &  & X & X &  & X & X &  & X & X &  & X & X &  & X & X & X & X & X & X & X & X & X & X \\
fin-2 & X &  &  &  &  & X & X & X & X & X &  & X & X &  & X & X &  & X & X &  & X & X &  &  \\

\bottomrule
\end{tabular}}
\caption{Checklist of LLMs that fail on specific templates. A failure is marked with an "X". An LLM is considered to fail a template if at least one of its evaluation units exhibits a failure rate of 60\% or higher.}
\label{tab:checklist}
\end{table*}
Table~\ref{tab:checklist} shows where all LLMs fail with \impact. We consider an LLM to fail on a specific template when the failure rate on at least one evaluation units is 40\% or higher.}

\blue{\section{Template Space}
\label{sec:appendixH}
We show the morphosyntactic features for the common templates for each language in Table~\ref{tab:template_space}.
\begin{table*}[h]
\centering
\renewcommand\cellalign{tl}

\resizebox{0.8\textwidth}{!}{\begin{tabular}{ccp{9.5cm}}
\toprule
\textbf{Template} & \textbf{Language} & \textbf{Features} \\
\midrule
\multirow{6}{*}{com-1} 
& eng & \makecell[l]{
\texttt{noun: \{plurality: [PL, SG]\}}\\
\texttt{verb: \{plurality: [PL, SG],}\\
\texttt{tense: [PST, PRS], mood: [IND]\}}} \\
& ara & \makecell[l]{
\texttt{noun: \{gender: [FEM, MASC],}
\texttt{plurality: [PL, DU, SG]\}}\\
\texttt{verb: \{gender: [FEM, MASC],}
\texttt{plurality: [PL, DU, SG],}\\
\texttt{aspect: [IPFV, PFV], mood: [IND]\}}} \\
& heb & \makecell[l]{
\texttt{noun: \{gender: [FEM, MASC],}
\texttt{plurality: [PL, SG]\}}\\
\texttt{verb: \{gender: [FEM, MASC],}
\texttt{plurality: [PL, SG],}\\
\texttt{tense: [PRS, PST], mood: [IND]\}}} \\
& rus & \makecell[l]{
\texttt{noun: \{gender: [FEM, MASC],}
\texttt{plurality: [PL, SG]\}}\\
\texttt{verb: \{plurality: [PL, SG],}
\texttt{tense: [PRS, PST],}\\
\texttt{mood: [IND], gender: [FEM, MASC]\}}} \\
& tur & \makecell[l]{
\texttt{noun: \{plurality: [SG, PL]\}}\\
\texttt{verb: \{plurality: [SG, PL],}
\texttt{tense: [PRS, PST],}\\
\texttt{mood: [IND], aspect: [PFV, PROG]\}}} \\
& fin & \makecell[l]{
\texttt{noun: \{plurality: [PL, SG]\}}\\
\texttt{verb: \{plurality: [PL, SG],}
\texttt{tense: [PST, PRS], mood: [IND]\}}} \\
\midrule
\multirow{6}{*}{com-2} 
& eng & \makecell[l]{
\texttt{noun: \{plurality: [PL, SG]\}}\\
\texttt{verb: \{plurality: [PL, SG],}
\texttt{mood: [IND]\}}} \\
& ara & \makecell[l]{
\texttt{noun: \{gender: [FEM, MASC],}
\texttt{plurality: [PL, DU, SG]\}}\\
\texttt{verb: \{gender: [FEM, MASC],}
\texttt{plurality: [PL, DU, SG],}\\
\texttt{mood: [IMP]\}}} \\
& heb & \makecell[l]{
\texttt{noun: \{gender: [FEM, MASC],}
\texttt{plurality: [PL, SG]\}}\\
\texttt{verb: \{gender: [FEM, MASC],}
\texttt{plurality: [PL, SG],}\\
\texttt{mood: [IND]\}}} \\
& rus & \makecell[l]{
\texttt{noun: \{gender: [FEM, MASC],}
\texttt{plurality: [PL, SG]\}}\\
\texttt{verb: \{plurality: [PL, SG],}
\texttt{mood: [IMP]\}}} \\
& tur & \makecell[l]{
\texttt{noun: \{plurality: [SG, PL]\}}\\
\texttt{verb: \{plurality: [SG, PL],}
\texttt{mood: [IMP]\}}} \\
& fin & \makecell[l]{
\texttt{noun: \{plurality: [PL, SG]\}}\\
\texttt{verb: \{plurality: [PL, SG],}
\texttt{mood: [IMP]\}}} \\

\midrule
\multirow{6}{*}{com-3} 
& eng & \makecell[l]{
\texttt{noun: \{plurality: [PL, SG]\}}}\\
& ara & \makecell[l]{
\texttt{noun: \{gender: [FEM, MASC],}\\
\texttt{plurality: [PL, DU, SG],}\\
\texttt{animacy: [HUM, NHUM]\}}\\
\texttt{adjective: \{gender: [FEM, MASC],}
\texttt{plurality: [PL, DU, SG],}
\texttt{case: [NOM]\}}
}\\
& heb & \makecell[l]{
\texttt{noun: \{gender: [FEM, MASC],}\\
\texttt{plurality: [PL, SG],}\\
\texttt{animacy: [HUM, NHUM]\}}\\
\texttt{adjective: \{gender: [FEM, MASC],}
\texttt{plurality: [PL, SG]\},}\\} \\
& rus & \makecell[l]{
\texttt{noun: \{gender: [NEUT, FEM, MASC],}
\texttt{plurality: [PL, SG],}\\
\texttt{case: [DAT, GEN, ESS, ACC, INS, NOM]\}}\\
\texttt{adjective: \{gender: [NEUT, FEM, MASC],}
\texttt{plurality: [PL, SG],}\\
\texttt{case: [DAT, GEN, ESS, ACC, INS, NOM]\}}}\\
& tur & \makecell[l]{
\texttt{noun: \{plurality: [SG, PL], case: [NOM],}\\
\texttt{animacy: [NHUM, HUM]\}}
\texttt{noun: \{plurality: [SG, PL],}\\
\texttt{animacy: [NHUM, HUM]\}}}\\
& fin & \makecell[l]{
\texttt{noun: \{plurality: [SG, PL]\},}\\
\texttt{\{case: ['PRT', 'GEN', 'ESS', 'TRANS', 'NOM']\}}\\
\texttt{adjective: \{plurality: [SG, PL]\},}\\
\texttt{\{case: ['PRT', 'GEN', 'ESS', 'TRANS', 'NOM']\}}\\
} \\
\bottomrule
\end{tabular}}
\caption{Morphosyntactic feature sets per language for common templates. We refer the reader to the \href{https://unimorph.github.io/doc/unimorph-schema.pdf}{Unimorph documentation} for abbreviations.}
\label{tab:template_space}
\end{table*}}

\blue{\section{Fusional vs Agglutinative}
\label{sec:appendixI}
A diagram showing the difference between fusional and agglutinative languages is shown in Figure~\ref{fig:agglutinative_vs_fusion}.
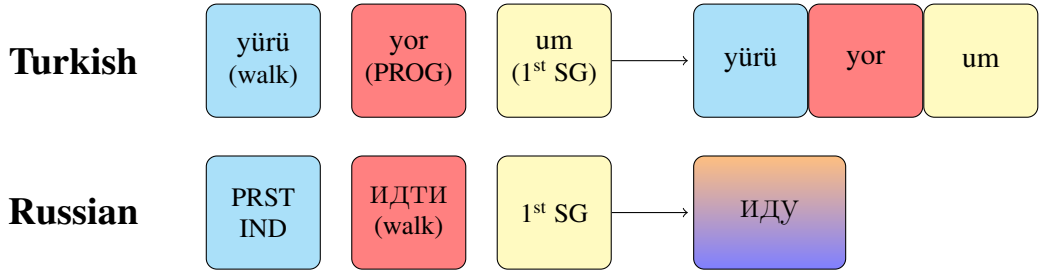
\begin{figure*}[ht]
    \centering
\begin{tikzpicture}[
    box/.style={rectangle, draw, rounded corners, minimum height=1.5cm, minimum width=1.5cm, align=center},
    arrow/.style={-Stealth, thick},
    font=\small
]

\node[align=right] (turkishLabel) at (-2.5,0) {\Large\bfseries Turkish};
\node[align=right] (russianLabel) at (-2.5,-2) {\Large\bfseries Russian};

\node[box, fill=cyan!30, minimum width=1.5cm] (pesu) at (0,0) {\normalsize yürü\\(walk)};
\node[box, fill=red!50, right=0.4cm of pesu, minimum width=1.5cm] (kir) {\normalsize yor\\(PROG)};
\node[box, fill=yellow!30, right=0.4cm of kir, minimum width=1.5cm] (en) {\normalsize um\\(1\textsuperscript{st} SG)};

\node[right=1cm of en] (arrow1) {}; 
\draw[->] (en) -- (arrow1);  

\node[box, fill=cyan!30, right=-0.2cm of arrow1, minimum width=1.5cm] (pesu2) {\normalsize yürü};
\node[box, fill=red!50, right=0cm of pesu2, minimum width=1.5cm] (kir2) {\normalsize yor};
\node[box, fill=yellow!30, right=0cm of kir2, minimum width=1.5cm] (en2) {\normalsize um};

\node[box, fill=cyan!30, below=0.5cm of pesu, minimum width=1.5cm] (ind) {\small PRST\\IND};
\node[box, fill=red!50, right=0.4cm of ind] (hablar) {\Large \foreignlanguage{russian}{идти}\\(walk)};
\node[box, fill=yellow!30, right=0.4cm of hablar] (yo) {1\textsuperscript{st} SG};

\node[right=1cm of yo] (arrow2) {};
\draw[->] (yo) -- (arrow2);

\node[box, top color=orange!50, bottom color=blue!50, right=-0.2cm of arrow2, minimum width=2cm] (hablo) {\Large  \foreignlanguage{russian}{иду}};

\end{tikzpicture}
    \caption{Turkish is a highly agglutinative language, meaning that each affix usually corresponds to a single grammatical function. For instance, in the phrase "I am walking", the verb root, progressive aspect marker, and first-person singular suffix are clearly distinguishable. In contrast, the other languages discussed are fusional: their affixes often blend multiple grammatical meanings into a single form, making morpheme boundaries less transparent, as illustrated in the example.}
    \label{fig:agglutinative_vs_fusion}
\end{figure*}
}

\blue{\section{A Note on Arabic Gender Polarity}
\label{sec:appendixJ}
While we refer to this phenomenon as gender reverse agreement for ease of exposition, we acknowledge that the phenomenon is more accurately characterized as morpheme-based and do not explore these details further here for simplicity~\citep{ling_a_00385}.
}

\blue{\section{Failing Examples}
\label{sec:appendixL}
Table~\ref{tab:failing_examples} shows some incorrect CoTs. Text in red shows where the reasoning fails.
\begin{table*}[]
    \centering
    \resizebox{\textwidth}{!}{\begin{tabular}{lcp{3cm}cp{7cm}}
      \textbf{Model}   & \textbf{Template} & 
      \textbf{Utterance} &
      \textbf{Gold}&
      \textbf{Reasoning}  \\
      \toprule
      Qwen3   & ara-com-3 & \textRL{ساعات قديمة} &Yes& 
Let's analyze the utterance **\textRL{ساعات قديمة}** step by step...

Step 2: Agreement in gender and number
- **\textRL{ساعات}** is a **feminine plural** noun.
- The adjective **\textRL{قديمة}** is **feminine singular**, not plural.
\textcolor{red}{So, if the noun is **\textRL{ساعات}** (feminine plural), the adjective should also be **feminine plural**, i.e., **\textRL{قديمات}** (qadīmat)...
Final Answer: No"}
       \\
      \midrule

      Aya   & tur-2 &  Nazlı \underline{\hspace{1cm}} &savunmuş& 
To fill in the blank in the utterance... considering the speaker did not witness the event personally, \textcolor{red}{we need to use the third-person singular past tense form of the verb in the ""duy-gör-bil"" (hear-see-know) construction.The correct form of ""savunmak"" in this context is ""savundu""...Final Answer: Nazlı savundu.}
       \\
      \midrule

      Phi4   & rus-com-3 &  \foreignlanguage{russian}{очаровательных ручки} &No& 
...let's break it down step by step:

1. Identify the words and their roles...

2. Check the agreement between the adjective and the noun...
3. Determine the correct form of the adjective...
\textcolor{red}{4. Check the case of the noun:
   - The noun is in the nominative case, which is the default case for the subject of a sentence. However, in this context, it seems to be used in the genitive case, which is often used to indicate possession or quantity.
5. Verify the overall structure:
   - The utterance is grammatically correct in Russian, as it follows the rules of agreement between the adjective and the noun, and the noun is in the genitive case.
Final Answer: Yes}
       \\
      \midrule
      GF2   & fin-1 & Pidän hallistä&No& Here's how to determine if the utterance is grammatically correct...
*   **Determining correctness**: Since ""halli"" is a consonant stem noun, the elative case is formed by adding ""-sta"" to the stem. Therefore, \textcolor{red}{""hallista"" is the correct elative form.
Final Answer: Yes} \\
      \midrule
      EuroLLM   & eng-com-1 &  Robert, Anthony, and Mark jumps &No& 
Step 1: Identify the subject of the sentence.
The subject of the sentence is ""Robert, Anthony, and Mark.""
Step 2: Determine the verb form required for the subject.
Since the subject consists of multiple nouns, the verb form should be plural.
Step 3: Check the verb form for the subject.
The verb form ""jumps"" is in the present tense and \textcolor{red}{is plural, making it grammatically correct for the subject ""Robert, Anthony, and Mark.""
Step 4: Determine if the sentence is grammatically correct.
The sentence ""Robert, Anthony, and Mark jumps"" is grammatically correct because it has a plural subject and a plural verb form in the present tense.
Final Answer: Yes}
       \\
      \midrule
      
    \end{tabular}}
    \caption{Table showing incorrect CoTs for some LLMs on various templates. \textcolor{red}{Text in red} highlights errors in the reasoning.}
    \label{tab:failing_examples}
\end{table*}}

\end{document}